\ifdefined\pdfobjcompresslevel
\fi
\PassOptionsToPackage{table,dvipsnames}{xcolor}
\documentclass[]{selfevolagent}
\usepackage{microtype}
\usepackage{amsfonts}
\usepackage{xcolor}
\definecolor{lavender}{RGB}{230,230,250}
\definecolor{softlavender}{RGB}{238, 223, 255}
\definecolor{selfevolagent}{RGB}{220,211,237}
\usepackage{wrapfig}
\usepackage{float}
\usepackage{amsmath}
\usepackage{amsthm}
\usepackage{subcaption}
\usepackage{makecell}
\usepackage{multirow}
\usepackage{algpseudocode}
\usepackage[linesnumbered,lined,boxed,commentsnumbered,ruled,longend]{algorithm2e}
\theoremstyle{plain}

\theoremstyle{definition}

\theoremstyle{remark}

\usepackage{enumitem}
\usepackage{pifont}
\usepackage[edges]{forest}
\usetikzlibrary{arrows.meta}
\usepackage{tikz}
\usepackage{graphicx}
\usepackage{booktabs}
\usepackage{enumitem}
\usepackage[most]{tcolorbox}
\usepackage{url}
\usepackage{tabularx}
\usepackage{fontawesome5}
\usepackage{svg}
\usepackage{xltabular}
\usepackage{caption}
\usepackage{titletoc}

\usetikzlibrary{
  positioning,
  calc,
  shapes.symbols,
  shapes.geometric,
  shapes.misc
}
\ifPDFTeX
\usepackage[T1]{fontenc}
\fi
\definecolor{selfevolagent_dark}{HTML}{37D2A6}
\definecolor{selfevolagent_light}{HTML}{9BE9D3}
\definecolor{selfevolagent_lighter}{HTML}{CDF4E9}
\definecolor{deltaaccent}{HTML}{23725F}
\newcommand{\deltacell}[1]{\cellcolor{selfevolagent_lighter!28}#1}
\newcommand{\deltalabel}{\deltacell{\textcolor{deltaaccent}{\textbf{$\Delta$ vs. init}}}}

\newcommand{\ghlink}[1]{\faIcon{github}\,\href{#1}{GitHub}}
\newcommand{\weblink}[1]{\faIcon{globe}\,\href{#1}{Website}}

\newcommand{\arbor}{\texttt{Arbor}}
\newcommand{\htr}{\textsc{HTR}}

\newcommand{\Mat}{\mathcal{M}}
\newcommand{\Obj}{\mathcal{O}}
\newcommand{\Eval}{\mathcal{E}}
\newcommand{\Tree}{\mathcal{T}\!\textit{ree}}
\newcommand{\dev}{\mathrm{dev}}
\newcommand{\test}{\mathrm{test}}

\definecolor{rootcolor}{RGB}{101, 45, 144}
\definecolor{catcolor}{RGB}{255, 192, 0}
\definecolor{subcatcolor}{RGB}{237, 125, 49}
\definecolor{papercolor}{RGB}{68, 114, 196}

\definecolor{prompttitle}{HTML}{2896FD}
\definecolor{promptbg}{HTML}{EDF2FB}
\definecolor{AppendixTOCSection}{HTML}{1E7F68}
\definecolor{AppendixTOCSubsection}{HTML}{2EA78A}
\definecolor{AppendixTOCSubsubsection}{HTML}{5E8F83}
\newcommand{\setupappendixtoc}{%
    \titlecontents{section}
        [2.6em]
        {\addvspace{0.35em}\small\bfseries\color{AppendixTOCSection}}
        {\contentslabel{2.3em}}
        {}
        {\titlerule*[0.45pc]{.}\contentspage}
    \titlecontents{subsection}
        [3.9em]
        {\small\color{AppendixTOCSubsection}}
        {\contentslabel{2.8em}}
        {}
        {\titlerule*[0.45pc]{.}\contentspage}
    \titlecontents{subsubsection}
        [6.2em]
        {\footnotesize\color{AppendixTOCSubsubsection}}
        {\contentslabel{3.5em}}
        {}
        {\titlerule*[0.45pc]{.}\contentspage}%
}

\newtcolorbox{promptbox}[1][]{
  colback=promptbg,
  colframe=prompttitle,
  coltitle=white,
  boxrule=0.5pt,
  breakable,
  arc=2pt,
  left=5pt, right=5pt, top=5pt, bottom=5pt,
  fontupper=\small,
  title=\textbf{System Prompt},
  #1
}
\newtcolorbox{toolbox}[1][]{
  colback=promptbg,
  colframe=prompttitle,
  coltitle=white,
  boxrule=0.5pt,
  breakable,
  arc=2pt,
  left=5pt, right=5pt, top=5pt, bottom=5pt,
  fontupper=\small,
  title=\textbf{Tool Definition},
  #1
}

\title{Toward Generalist Autonomous Research via Hypothesis-Tree Refinement}

\author[1,\dagger,\ddagger]{Jiajie Jin}
\author[1,\dagger]{Yuyang Hu}
\author[2]{Kai Qiu}
\author[2]{Qi Dai}
\author[2]{Chong Luo}
\author[1]{Guanting Dong}
\author[1]{Xiaoxi Li}
\author[1]{Tong Zhao}
\author[2]{Xiaolong Ma}
\author[2]{Gongrui Zhang}
\author[2]{Zhirong Wu}
\author[2]{Bei Liu}
\author[2]{Zhengyuan Yang}
\author[2]{Linjie Li}
\author[2]{Lijuan Wang}
\author[1]{Hongjin Qian}
\author[1]{Yutao Zhu}
\author[1,*]{Zhicheng Dou}

\affiliation[1]{Gaoling School of Artificial Intelligence, Renmin University of China}
\affiliation[2]{Microsoft Research}
\contribution[\dagger]{Equal contribution}
\contribution[\ddagger]{Work done during an internship at MSRA}
\contribution[*]{Corresponding author}

\abstract{%
Scientific progress depends on a repeated loop of exploration, experimentation, and abstraction. Researchers test candidate directions, interpret the evidence, and carry the resulting lessons into later attempts. We study how an AI agent can run this loop autonomously over long horizons. We introduce \arbor{}, a general framework for autonomous research that combines a long-lived coordinator, short-lived executors, and Hypothesis Tree Refinement (HTR), a persistent tree that links hypotheses, artifacts, evidence, and distilled insights across time. The coordinator manages global research strategy over the tree, while executors implement and test individual hypotheses in isolated worktrees. As results return, \arbor{} updates the tree, propagates reusable lessons, refines the search frontier, and admits verified improvements. This design turns autonomous research from a sequence of local attempts into a cumulative process in which strategy, execution, and evidence are carried across time.
We evaluate \arbor{} under \emph{Autonomous Optimization} (AO), an operational setting where an agent improves an initial research artifact through iterative experimentation without step-level human supervision. Across six real research tasks in model training, harness engineering, and data synthesis, \arbor{} achieves the best held-out result on all six tasks, attaining more than $2.5\times$ the average relative held-out gain of Codex and Claude Code under the same task interface and resource budget. On MLE-Bench Lite, \arbor{} reaches 86.36\% Any Medal with GPT-5.5, the strongest result in our comparison. 

\textit{Note: This is a living technical report for an ongoing project. We will continue to refine and expand the scope of evaluation as the project evolves, and will update this report accordingly.}
}
\metadata[\faEnvelope\ Contact]{jinjiajie@ruc.edu.cn, dou@ruc.edu.cn}

\metadata[{\raisebox{-0.2ex}{\includegraphics[height=1.1em]{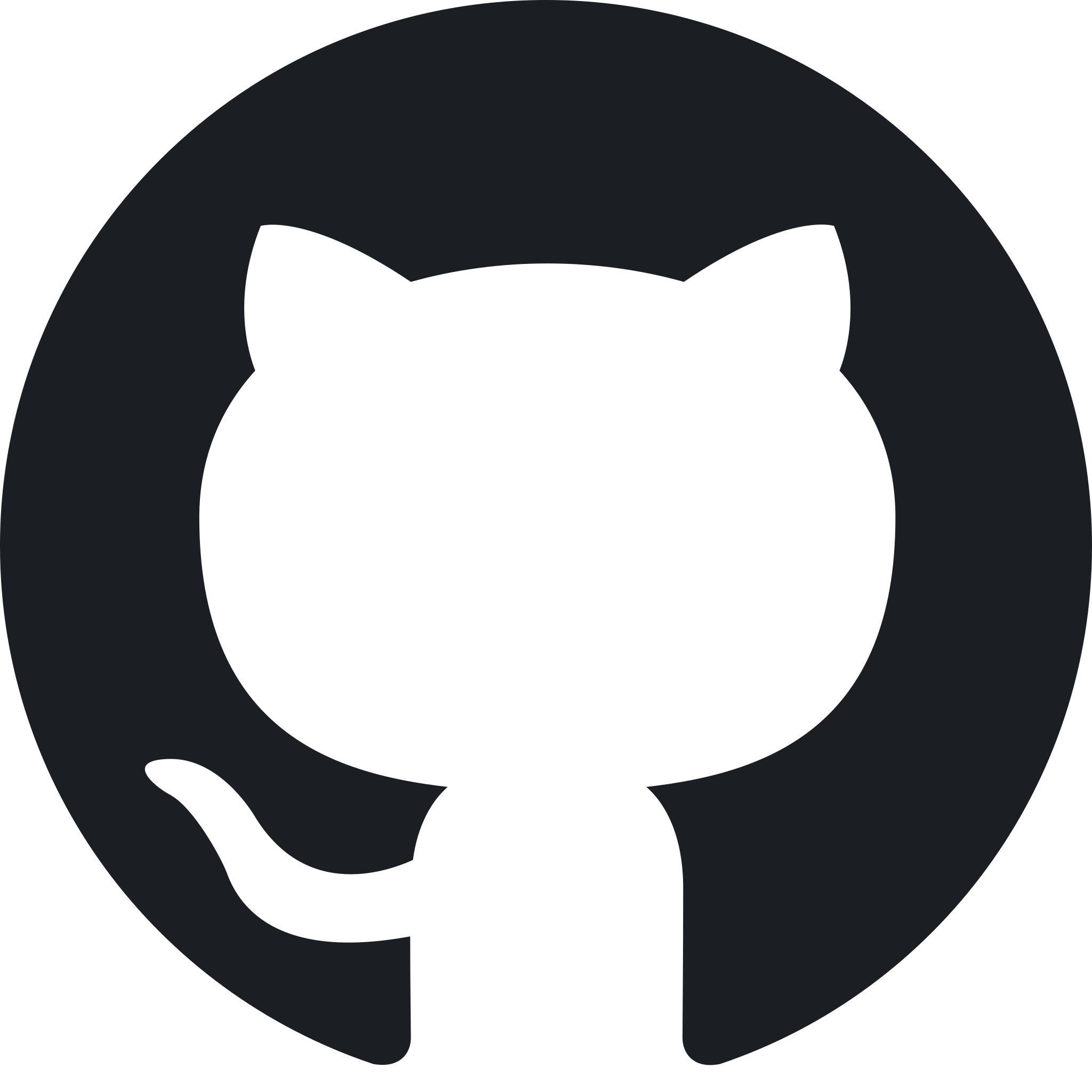}}\ Code}]{\url{https://github.com/RUC-NLPIR/Arbor}}

\begin{document}

 \maketitle

\vspace{-0.3cm}
 \begin{figure}[H]
    \centering
  \includegraphics[width=\linewidth]{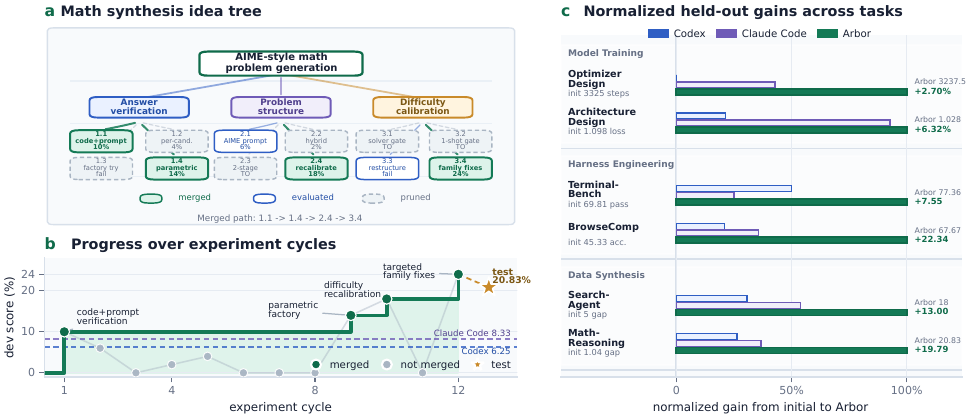}
  \caption{\arbor{} at a glance. \textbf{(a)} Hypothesis tree and \textbf{(b)}
  development score from one Math-Reasoning Data Synthesis run. \textbf{(c)}
  Normalized held-out gains across all tasks.}
    \label{fig:teaser}
\end{figure}

\section{Introduction}\label{sec:intro}

Scientific research is a central form of long-horizon human intelligence~\citep{aiscientist_long,aiscientistsurvey}. Its difficulty lies not only in solving isolated problems, but in sustaining progress across uncertain hypotheses, costly experiments, failed attempts, and delayed feedback. A researcher must maintain an evolving understanding of the problem so that each attempt can reshape what should be tried next~\citep{memorysurvey}. Recent LLM agents can now edit code, call tools, retrieve information, and run experiments for extended periods~\citep{react,toolformer,toolllm}, and systems such as Codex~\citep{codex}, Claude Code~\citep{claudecode}, and OpenHands~\citep{openhands} make sustained progress in real codebases, making autonomous research an increasingly concrete systems problem. Yet longer execution alone does not guarantee research progress. The open challenge is how an agent can maintain a research state that turns many local attempts into cumulative hypothesis refinement and verified artifact improvement.

We formalize this problem as \emph{Autonomous Optimization} (AO), which captures the core operational form of autonomous research. In AO, an agent begins with an initial artifact and a research objective, then improves the artifact through experimental feedback without step-level human supervision. This setting is difficult because research feedback is delayed, experiments can be expensive, and failed attempts often contain information that should guide later search. As the horizon grows, an agent that treats each trial as an independent local attempt loses the structure of the research process. Effective AO therefore requires a persistent research state that records what has been tried, what evidence was obtained, and how each result changes the space of future hypotheses.

Despite recent progress, current agent systems still do not provide a general framework for running autonomous research over long horizons~\citep{metaharness,autoharness}. General coding agents can edit code, invoke tools, and run experiments for many hours, but their autonomy is mostly expressed as persistent task execution. Scientific-agent systems move closer to research automation, yet many still follow predefined workflows or revise a single line of work at a time~\citep{aiscientist,agentlaboratory,aflow}. They therefore lack the mechanism that makes human research cumulative: the ability to maintain competing directions, test them through concrete experiments~\citep{treeofthoughts,lats}, interpret both successes and failures~\citep{memorysurvey}, and let those lessons reshape later exploration. For AO, the key challenge is to build this mechanism into the agent system itself, so that long-running experimentation becomes a self-directed research process rather than an extended sequence of local attempts.

We argue that a general AO system should automate the long-horizon work that a human researcher normally performs during iterative research. Starting from an open objective, it should form research directions, test them through concrete artifact changes, and turn the resulting evidence into memory that shapes later exploration. Progress should not depend on a human repeatedly choosing the next attempt or interpreting what previous trials mean. Instead, the system needs a framework that keeps directions, experiments, artifacts, results, and failures connected across time, turning autonomous research into a persistent cycle of exploration and verified improvement.

We introduce \textbf{Arbor}, a general framework and open-source research system for AO. \arbor{} separates autonomous research into a long-lived coordinator and short-lived executors. The coordinator owns the global research state and decides how the search frontier should evolve, while each executor tests one hypothesis in an isolated worktree and returns structured evidence. \arbor{} makes this two-level process cumulative through Hypothesis Tree Refinement (HTR). HTR represents the research process as a persistent tree in which each node binds a hypothesis, the artifact version that realizes it, the experimental evidence it produces, and the distilled insight that should shape later decisions. When executor results return, \arbor{} writes evidence back to the executed nodes, abstracts local findings upward, and uses the updated tree to decide which directions to expand, prune, or merge, promoting a candidate to the current best only when it improves a held-out evaluation. The tree therefore acts as the operational research state of the system: it is simultaneously the search frontier, the memory of past attempts, and the audit trail for verified artifact improvement.

To evaluate \arbor{}, we construct six AO tasks from real research settings across model training~\citep{nanogptbench,autoresearch}, harness engineering~\citep{terminalbench,browsecomp}, and data synthesis. Each task specifies an initial artifact, a natural-language objective, a task-native metric, and a development/test protocol that separates exploratory feedback from final scoring. \arbor{} achieves the best held-out result on all six tasks, with more than $2.5\times$ the average relative gain of Codex and Claude Code under the same task interface and resource budget. On MLE-Bench Lite~\citep{mlebench}, \arbor{} further reaches 86.36\% Any Medal with GPT-5.5, the strongest result in our comparison. Ablations, backbone studies, transfer experiments, and cost analyses show that these gains come from \arbor{}'s evidence-structured research process: hypotheses remain grounded in executable artifacts, local findings become reusable insights, and later decisions are made over an explicit research state.

Our contributions are summarized as follows:

\begin{itemize}
\item We formulate \emph{Autonomous Optimization} (AO) as a class of long-horizon research tasks in which an agent must iteratively improve an artifact under a fixed objective and evaluator without step-level human supervision.
\item We introduce \textbf{Arbor}, a general framework for AO that organizes research through \emph{Hypothesis Tree Refinement} (HTR), pairing a persistent coordinator with isolated executors so that hypotheses, artifact versions, experimental evidence, and distilled insights accumulate into an auditable research state; we release it as an open-source research system.
\item We construct six AO tasks from real research settings and show, together with MLE-Bench Lite, that \arbor{} delivers the strongest held-out gains and that persistent hypothesis management and insight propagation are the key drivers of its performance.
\end{itemize}

\section{Related Work}\label{sec:related}
\subsection{Autonomous Research Agent}
LLM-based automated research systems first appeared as end-to-end pipelines. The AI Scientist~\citep{aiscientist} connected idea generation, implementation, execution, result interpretation and paper writing in a mostly automated loop, while Agent Laboratory~\citep{agentlaboratory} organized similar stages as a human-supervised multi-agent research-assistant workflow. The next wave made the search process more explicit: AIDE~\citep{aide} explored ML engineering as iterative code search, AI Scientist-v2~\citep{aiscientistv2} introduced agentic tree search over research plans and experiments, and multi-agent systems such as AI-Researcher, R\&D-Agent~\citep{rdagent} and Loongflow~\citep{loongflow} refined the literature-to-experiment loop through more specialized roles and summarization mechanisms. In parallel, FunSearch~\citep{funsearch} and AlphaEvolve~\citep{alphaevolve} treated LLMs as program mutation operators selected by executable fitness signals, and SciMaster~\citep{scimaster} broadened automated research from ML experimentation to general-purpose scientific reasoning with tool-augmented, breadth-and-depth search.

Recent systems expand the object of search itself. MARS~\citep{mars} modularizes automated AI research into reflective components; AutoHarness~\citep{autoharness}, Meta-Harness~\citep{metaharness} and AHE~\citep{ahe} search or evolve the code harness surrounding an agent; and DataMaster~\citep{datamaster} moves the search target to data, using DataTree, Data Pool and Global Memory to organize autonomous data discovery and validation. Arbor instead stores research state in a persistent hypothesis tree. A long-running coordinator expands and updates the tree, while short-lived executors implement individual hypotheses in isolated git worktrees. This design makes hypotheses, failures, evidence and merge decisions auditable. It also addresses gaps noted by recent surveys and sandbagging studies: weak evidence preservation, loose dev/test discipline and silent metric chasing~\citep{aiscientistsurvey,gasteiger2025automated}.

\subsection{Long-Horizon Agent}
As language agents have improved, the key question has shifted from whether they can complete isolated tool-use episodes to how long they can remain coherent on real tasks~\citep{longcontextsurvey,longhorizonexecution,agentfugue}. Early systems such as Reflexion~\citep{reflexion} and Generative Agents~\citep{generativeagent} extended single-run behavior with natural-language memories or reflections across trials, making experience accumulation part of the agent loop. Later human-calibrated evaluations made this horizon measurable: some tasks compare agents with human time-to-complete~\citep{rebench,hcast,metr}, while others~\citep{nanogptspeedrunning,posttrainbench} show that agents still struggle to preserve and reuse evidence across long optimization histories, even when the environment provides executable feedback.

Recent approaches therefore increasingly treat long-horizon agency as a problem of externalized state organization rather than only prompt design. Some systems organize prior experience into persistent context, using curated playbooks~\citep{ace}, state-adaptive trajectory retrieval~\citep{sam}, or cognitive caches~\citep{mlmaster2} so that later decisions can draw on earlier failures and successes. Others push the same idea into the scaffold around the model, coordinating agents through persistent workspaces~\citep{aiscientist_long}, recursively modifying agent code~\citep{darwinmachine}, or evolving harnesses that shape tool use and execution constraints~\citep{autoharness}. Arbor follows this state-externalization trend, but its persistent object is specifically a research tree: each node binds a hypothesis, implementation branch, result, score, related work and learned insight, so progress accumulates through branch expansion, insight backpropagation, merge decisions and pruning rather than through an ever-growing context window.

\subsection{Benchmark for Autonomous Research}
Research-agent benchmarks have progressed along several complementary directions. MLAgentBench~\citep{mlagentbench}, MLE-bench~\citep{mlebench} and MLE-Dojo~\citep{mledojo} evaluate agents on executable ML-engineering workflows with objective task metrics. ScienceAgentBench~\citep{scienceagentbench}, PaperBench~\citep{paperbench}, and FrontierScience~\citep{frontierscience}focus on programmatic discovery, or paper reproduction under expert-designed questions and rubrics. RE-Bench~\citep{rebench}, HCAST~\citep{hcast}, AlgoTune~\citep{algotune}, NanoGPT speedrunning~\citep{nanogptspeedrunning}, PostTrainBench~\citep{posttrainbench} and Frontier-Eng~\citep{frontiereng} further emphasize long-horizon engineering, human-calibrated difficulty, executable feedback and iterative self-improvement.

These benchmarks measure important outcomes, but many evaluations still use incomplete settings: some lack a clear dev/test split, making iterative search prone to overfitting, while others rely on one or two task types and leave generality under-tested. Arbor therefore evaluates across multiple tasks and enforces a stricter protocol: data and evaluation harnesses are immutable, the dev set supports iterative search, the test set is reserved for merge or final validation, and each experiment is tied to branch-level artifacts and hypothesis-tree records.

\begin{figure}[t]
    \centering
  \includegraphics[width=\linewidth]{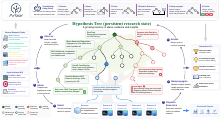}
    \caption{Overall framework of \arbor{}. A persistent coordinator maintains
    the research state as a hypothesis tree, iteratively exploring ideas,
    dispatching executors to implement them, and using evaluation feedback to
    refine the tree and update the current best artifact.}
    \label{fig:architecture}
\end{figure}

\section{Task Formulation}\label{sec:task}\label{sec:ao-interface}
We model auto-research as an instance of \emph{Autonomous Optimization}
(AO), which can be represented as a tuple
\[
\mathcal{P}=(\Mat_0,\Obj,\Eval_{\dev},\Eval_{\test}).
\]
The material $\Mat_0$ is the mutable artifact the agent may inspect and
modify, typically a codebase together with its associated data. The
objective $\Obj$ specifies what it means for a modified material $\Mat'$
to be better, for example a metric direction defined over the artifact's
output. The two evaluators instantiate the same objective on different
evidence: $\Eval_{\dev}$ returns feedback the agent may freely use
during search, while the held-out $\Eval_{\test}$ measures whether the
dev-driven improvement transfers beyond the feedback used for
exploration. Let $S_{\dev}(\Mat')$ and $S_{\test}(\Mat')$ denote the
scalar scores returned by these two evaluators under the metric
direction specified by $\Obj$, so that larger values are better; a
candidate that exploits idiosyncrasies of the dev split may improve
$S_{\dev}$ but is not a successful AO solution unless the gain also
transfers to $S_{\test}$.

During a run, the agent adaptively generates, implements, and evaluates
candidate materials using $\Eval_{\dev}$. Let $\mathcal{A}$ be the set
of candidates produced. The artifact-level goal is to return
\[
\Mat^\star = \arg\max_{\Mat' \in \mathcal{A}} S_{\test}(\Mat'),
\]
subject to the constraint that hypotheses and implementation decisions
are made without using $\Eval_{\test}$ as an exploration oracle.

\section{The \arbor{} Framework}\label{sec:framework}

\subsection{Overview}\label{sec:framework-overview}

We propose \arbor{}, a general framework for autonomous research under the AO interface defined in Section~\ref{sec:ao-interface}. AO differs from ordinary agentic tool use in that the target is not a single response or code patch, but a sustained research trajectory. An agent must propose hypotheses, materialize them as artifact changes, interpret experimental feedback, and decide which directions should be refined, merged, or abandoned. The central design problem is therefore how to convert many transient trials into cumulative research progress.

This problem imposes three requirements on the system design:

\begin{itemize}[leftmargin=*]

    \item \textbf{Branching with coherence.} Research exploration must branch because multiple competing hypotheses may be plausible at the same time. However, unrestricted branching can degenerate into an unstructured log of attempts. The system must therefore maintain a frontier in which competing directions coexist while remaining organized, comparable, and actionable.

    \item \textbf{Global strategy with local execution.} Strategic decisions depend on evidence accumulated across the whole run, whereas implementing a single hypothesis requires short-horizon code editing, debugging, and evaluation. These two levels should be separated so that low-level execution traces do not obscure the global research state, and experimental outcomes remain attributable to the hypotheses that produced them.

    \item \textbf{Exploration with held-out admission.} Development feedback should guide hypothesis search, but artifact-level progress should be admitted only when it transfers beyond the feedback used during exploration. The system must therefore distinguish exploratory improvement on $E_{\mathrm{dev}}$ from verified improvement under the held-out evaluator $E_{\mathrm{test}}$.

\end{itemize}

\arbor{} addresses these requirements through Hypothesis Tree Refinement (HTR), as illustrated in Figure~\ref{fig:architecture}. Its central state is a persistent hypothesis tree whose nodes bind together a research hypothesis, the artifact version that realizes it, the evaluation evidence it produces, and the distilled insight that should influence later decisions. A long-lived coordinator maintains this tree as the global research state: it observes the current frontier, proposes refinements, selects promising leaves, integrates returned evidence, propagates insights upward, and decides whether to continue, prune, or merge a candidate branch. Short-lived executors test selected hypotheses in isolated worktrees and return compact reports containing scores, factual results, distilled insights, and artifact references. A held-out merge gate promotes a candidate to the current best artifact only when its improvement transfers beyond the development evaluator. In this way, \arbor{} organizes AO as evidence-structured refinement over a durable research state rather than repeated local trial-and-error. Section~\ref{sec:framework-htr} describes the hypothesis-tree representation, and Section~\ref{sec:framework-loop} presents the coordinator--executor loop that maintains it.

\subsection{Hypothesis Tree as Research State}
\label{sec:framework-htr}
The center of Figure~\ref{fig:architecture} shows the hypothesis tree that serves as \arbor{}'s persistent research state. In AO, the intermediate state is not only the latest artifact or its evaluation score, but also the structure of exploration: which hypotheses have been considered, how they relate to one another, what evidence they produced, and what lessons should constrain future trials. A tree is a natural representation for this state because it preserves both the branching structure of research exploration and the abstraction hierarchy from broad directions to executable interventions.

Let $\mathcal{T}=(\mathcal{V}, \mathcal{E})$ denote a rooted hypothesis tree with root node $n_0$. Each node $n\in\mathcal{V}$ is a research unit
\[
n = \langle h_n, \iota_n, \mu_n \rangle,
\]
where the three fields separate the semantic content of a hypothesis, the reusable evidence derived from it, and the executable record that grounds it:

\begin{itemize}
    \item \textbf{Hypothesis $h_n$.} 
    The hypothesis describes a verifiable or falsifiable claim about how the material should be changed to improve the objective. Its granularity depends on the node depth: nodes close to the root describe broad research directions, while deeper nodes specify concrete interventions that can be implemented and evaluated by an executor. This allows \arbor{} to organize exploration as progressive refinement rather than as a flat sequence of independent trials.

    \item \textbf{Insight $\iota_n$.} 
    The insight stores the reusable interpretation of evidence associated with the hypothesis. For an executed leaf, it summarizes what was tried, what happened, and why the result supports, weakens, or constrains the hypothesis. For an internal node, it abstracts over the insights of its children and summarizes the current understanding of that research direction. Thus, $\iota_n$ is not an execution transcript, but a compact semantic memory for later hypothesis generation and selection.

    \item \textbf{Metadata $\mu_n$.} 
    The metadata connects the semantic hypothesis to executable evidence. It includes the node status, development score when available, factual result record, implementation reference such as a git branch or commit, and optional background evidence. The material itself is not duplicated in the tree; instead, the tree stores references to external artifact states produced in isolated worktrees. This keeps the research state compact while ensuring that each hypothesis remains grounded in a verifiable implementation.
\end{itemize}

The tree separates internal direction nodes from executable leaf nodes. Internal nodes maintain abstract research directions and accumulated lessons, whereas leaves represent candidate interventions that can be dispatched for implementation and evaluation. After a leaf is executed, its score, result, artifact reference, and distilled insight are written back to the corresponding node. The insight is then propagated upward by updating the ancestors along the path to the root. Through this abstraction process, local experimental outcomes become direction-level lessons and eventually contribute to a compact global understanding of the run.

In this way, the hypothesis tree serves three roles simultaneously. It is a search frontier that records which directions remain active, validated, or pruned; a long-term memory that stores reusable evidence from both successes and failures; and an auditable research record that links each artifact change to the hypothesis and evidence that motivated it. This persistent state provides the substrate on which the coordinator can make strategic decisions across long-horizon autonomous optimization.

\begin{algorithm}[!t]
\footnotesize
\DontPrintSemicolon
\SetKwInOut{Input}{Input}\SetKwInOut{Output}{Output}
\SetKwProg{Fn}{Procedure}{:}{}
\SetKwFunction{Exec}{Executor}
\Input{$\mathcal{P}=(\Mat_0,\Obj,\Eval_{\dev},\Eval_{\test})$, budget $B$, branching $k$}
\Output{best artifact $\Mat^\star$ and hypothesis tree $\Tree$}
init $\Tree=(\{n_0\},\emptyset)$, $b_{n_0}\leftarrow\Mat_0$, $\Mat_{\mathrm{best}}\leftarrow\Mat_0$\;
\While{$B$ left $\wedge$ pending leaves exist}{
  $\mathcal{V}\leftarrow\textsc{Observe}(\Tree,\Mat_{\mathrm{best}})$
    \tcp*{\textsc{Observe}: shape, root insight, pruned/validated lessons}
  $p\leftarrow$ choose parent under $\mathcal{V}$;\quad
  attach $k$ pending children $\{n^{(i)}:h^{(i)}\}\leftarrow\textsc{Ideate}(p,\mathcal{V})$
    \tcp*{\textsc{Ideate}}
  $L\leftarrow$ pending leaves under $\textsc{Select}(\mathcal{V})$
    \tcp*{\textsc{Select}: frontier control}
  $\{(s_n,r_n,\iota_n,b_n)\}_{n\in L}\leftarrow\textbf{parallel }\Exec(h_n,\iota_{\mathrm{anc}(n)},\Mat_{\mathrm{best}})$
    \tcp*{\textsc{Dispatch}}
  \ForEach{$n\in L$, $a\in\textsf{path}(n_0\to n)$}{
    write back $(s_n,r_n,\iota_n,b_n)$;\quad
    $\iota_a\leftarrow\textsc{Abstract}(\{\iota_c\}_{c\in\textsf{ch}(a)})$
      \tcp*{\textsc{Backpropagate}}
  }
  $n^\dagger\leftarrow\arg\max_{n\in L}s_n$
    \tcp*{\textsc{Decide}: held-out merge gate, then prune}
  \lIf{$\Obj(\Eval_{\test}(b_{n^\dagger}))>\Obj(\Eval_{\test}(\Mat_{\mathrm{best}}))$}
      {$\Mat_{\mathrm{best}}\leftarrow\textsf{merge}(b_{n^\dagger})$}
  prune subtrees falsified by $\{\iota_n\}_{n\in L}$;\quad persist $\Tree$\;
}
\Return{$\Mat^\star\leftarrow\Mat_{\mathrm{best}}$, $\Tree$}\;
\Fn{\Exec{$h_n,\iota_{\mathrm{anc}(n)},\Mat_{\mathrm{best}}$}}{
  fresh worktree $W_n\leftarrow\Mat_{\mathrm{best}}$\;
  \Repeat{run ok $\wedge$ $h_n$-path exercised, or cap reached}{
    $\Delta\leftarrow\textsf{Implement}(h_n,\iota_{\mathrm{anc}(n)},W_n)$;\quad
    $(s_n,r_n)\leftarrow\Eval_{\dev}(\textsf{apply}(\Delta,W_n))$
      \tcp*{repair $\Delta$ only; $h_n$ is fixed}
  }
  \Return{$(s_n,\,r_n,\,\textsc{Distill}(h_n,\Delta,r_n),\,\textsf{commit}(W_n))$}\;
}
\caption{Hypothesis Tree Refinement (HTR). Coordinator owns $\Tree$; \textsc{Executor} owns one worktree.}
\label{alg:htr}
\end{algorithm}

\subsection{Hypothesis Tree Refinement}
\label{sec:framework-loop}
\label{sec:framework-roles}
To maintain the tree over a long-horizon AO run, \arbor{} separates global
frontier control from local experimental execution. A persistent
\emph{coordinator} owns the shared tree and decides where to expand, which
evidence to trust, which directions to prune, and when a candidate should be
merged. Short-lived \emph{executors} are invoked only to test individual
hypotheses: each executor receives one tree node, materializes the corresponding
intervention in an isolated git worktree, evaluates it, and returns structured
evidence to the coordinator.

During the research process, the coordinator sees the whole research
frontier but does not directly perform every low-level implementation step; the
executor performs grounded engineering work but does not modify the shared tree
or redirect the search objective. As a result, exploratory code changes remain
isolated until they pass the merge gate, while the tree records only
decision-relevant evidence: scores, factual outcomes, artifact references, and
distilled insights. This boundary allows \arbor{} to turn transient execution
traces into a persistent research state without reducing the tree to a raw log
of tool calls.

\subsubsection{Coordinator: Evidence-Aware Frontier Control}
\label{sec:framework-loop-policy}
The coordinator updates $\Tree$ through a repeated six-step procedure:
\textsc{Observe}, \textsc{Ideate}, \textsc{Select}, \textsc{Dispatch},
\textsc{Backpropagate}, and \textsc{Decide}. Each step operates on the tree
through a narrow interface for adding nodes, dispatching executors, updating
node evidence, propagating insights, pruning subtrees, and merging verified
branches. The key point is that the LLM policy chooses how to interpret the
research state, while all durable state changes are expressed as controlled
mutations of the hypothesis tree.

\textbf{\textsc{Observe}.}
At the beginning of each cycle, the coordinator re-grounds itself in the current
research state by reading a structured projection of $\Tree$, including active
frontier nodes, recently returned evidence, ancestor insights, and the current
best artifact $\Mat_{\mathrm{best}}$. This step makes the tree the authoritative
state after context compression and prevents the coordinator from relying on a
lossy conversational history.

\textbf{\textsc{Ideate}.}
The coordinator selects a parent node and proposes a small set of child
hypotheses beneath it. Each child represents a refinement, alternative, or
correction of the parent hypothesis and is initialized as a pending node. Unlike
free-form brainstorming, ideation is conditioned on accumulated tree evidence:
validated insights provide assumptions to build on, pruned nodes provide
negative constraints, and recent executor reports suggest which interventions
are feasible or under-tested.

\textbf{\textsc{Select}.}
The coordinator chooses pending nodes to execute next. Selection balances the
expected utility of a hypothesis with the evidence already accumulated around
its ancestors and siblings. A direction may be selected because it has strong
prior evidence, because its siblings expose an unresolved ambiguity, or because
its failure would clarify an important assumption. Thus selection is not merely
score maximization; it is frontier control under partial and delayed feedback.

\textbf{\textsc{Dispatch}.}
Selected hypotheses are dispatched to independent executors. Each executor
materializes its assigned hypothesis in a fresh worktree, evaluates the modified
artifact on $\Eval_{\dev}$, and returns a compact report containing the dev
score, factual result, distilled insight, and branch reference. Parallel
execution of sibling hypotheses provides comparative evidence within the same
research direction, which is useful for later pruning and abstraction.

\textbf{\textsc{Backpropagate}.}
When executor reports return, the coordinator writes their evidence into the
corresponding leaf nodes and updates insights along the path to the root. The
propagated signal is not only a scalar score. It also includes causal
attributions, applicability conditions, and reusable lessons extracted from the
experiment. A leaf-level observation such as a data-interface mismatch can
therefore become a direction-level constraint, and eventually a global prior
that shapes future ideation.

\textbf{\textsc{Decide}.}
After the tree absorbs the new evidence, the coordinator decides whether to
continue expanding a direction, prune a falsified subtree, stop the run, or
attempt to merge a candidate branch. Promotion is guarded by a held-out merge
gate: the candidate is evaluated on $\Eval_{\test}$ in a fresh worktree and is
merged into $\Mat_{\mathrm{best}}$ only if it improves over the current best
under the objective $\Obj$. This gate separates exploratory success on
$\Eval_{\dev}$ from verified artifact-level progress.

\subsubsection{Executor: Hypothesis-Bound Experimentation}
\label{sec:framework-executor}
An executor implements one local experiment for one assigned hypothesis. Given a node $n$, it receives the hypothesis $h_n$, relevant ancestor insights, the
current best artifact, and the development evaluator $\Eval_{\dev}$. It then
creates an isolated worktree, applies the minimal intervention needed to realize
$h_n$, runs the evaluator, inspects failures or inactive code paths, and repairs
its own implementation when necessary. This local loop may involve multiple
edits and reruns, but it remains bound to the assigned hypothesis.

The executor returns exactly the evidence consumed by the coordinator's tree
interface: a comparable dev score for selection, a factual result for future
ideation, a distilled insight for backpropagation, and a branch reference for
held-out verification. This contract is important. If an executor were allowed
to change the hypothesis when the metric stalls, the returned score would no
longer be evidence about the assigned node, and ancestor-level insights would
become difficult to interpret. By keeping executors hypothesis-bound, \arbor{}
keeps local engineering flexibility while preserving the semantic meaning of
tree updates. Algorithm~\ref{alg:htr} summarizes the full HTR procedure.

\FloatBarrier

\section{Experiments}
\label{sec:experiments}

\subsection{AO Task Suite}
\label{sec:ao-task-suite}

To test whether \arbor{} can improve real research artifacts, we first construct
several AO tasks from actual research tasks. Each task consists of an initial
material $\Mat_0$, a natural language objective $\Obj$, an executable
development evaluator $\Eval_{\dev}$, a held-out test evaluator $\Eval_{\test}$,
and a task-native metric. Table~\ref{tab:ao-tasks} gives a compact summary; the
task details are described below.

\paragraph{Model training.}
The model-training tasks evaluate whether an agent can improve training
algorithms under expensive experimental feedback. In \textbf{Optimizer Design},
we use NanoGPT-Bench~\citep{nanogptbench}, a benchmark for accelerating
NanoGPT training. The initial material is the official tuned Muon optimizer
baseline distributed with NanoGPT-Bench, and the objective is to reach the
target NanoGPT validation loss in as few optimization steps as possible. The
development evaluator uses the standard NanoGPT-Bench task during search, while
the test evaluator reruns the selected optimizer with two held-out random seeds
and reports the average number of steps. In \textbf{Architecture Design}, we use
the \texttt{autoresearch} benchmark~\citep{autoresearch}. The agent modifies a
given LLM training codebase, with the goal of obtaining a lower final loss under
a fixed time budget. The test evaluator again averages two held-out random-seed
runs.

\paragraph{Harness engineering.}
The harness-engineering tasks evaluate whether an agent can improve the control
logic around another agent. In \textbf{Terminal-Bench 2.0}, the initial material
is the standard official terminal-agent codebase for Terminal-Bench
2.0~\citep{terminalbench}, and the objective is to improve pass rate on
terminal-based code and shell tasks. We stratify the 89 tasks by difficulty into
36 development tasks and 53 held-out test tasks, rather than optimizing on the
full benchmark. In \textbf{BrowseComp}, the initial material is our standard
minimal ReAct-style search harness~\citep{react,browsecomp}. The objective is to
improve answer accuracy on browsing questions; the development and test sets are
50 and 300 non-overlapping BrowseComp questions, respectively.

\paragraph{Data synthesis.}
The data-synthesis tasks evaluate whether an agent can improve a generation
pipeline whose output is judged by downstream model behavior. In
\textbf{Search-Agent Data Synthesis}, the initial material is a hand-designed
pipeline for generating search-agent questions from seed knowledge. Development
uses 50 seed items and test uses 100 disjoint seed items. In
\textbf{Math-Reasoning Data Synthesis}, the initial material is a hand-designed
pipeline for generating AIME-style reasoning problems; development generates 50
problems with 10 seed and test generates 96 problems with 12 seed. Both tasks are scored by the mean
$\mathrm{pass@4}-\mathrm{pass@1}$ gap under a strong GPT-5.5-based ReAct
evaluator. This metric rewards problems that are not solved immediately but can
be solved with additional attempts.

\begin{table}[t]
\centering
\small
\setlength{\tabcolsep}{3.0pt}
\renewcommand{\arraystretch}{1.08}
\caption{Compact summary of the AO tasks. Detailed task definitions are given in
Section~\ref{sec:ao-task-suite}.}
\label{tab:ao-tasks}
\begin{tabularx}{\linewidth}{>{\raggedright\arraybackslash}p{0.17\linewidth}>{\raggedright\arraybackslash}p{0.24\linewidth}>{\raggedright\arraybackslash}X>{\raggedright\arraybackslash}p{0.20\linewidth}}
\toprule
Type & Task & Initial material & Metric and split \\
\midrule
Model Training & Optimizer Design & NanoGPT-Bench; tuned Muon baseline & Steps to
target loss ($\downarrow$); test averages two seeds \\
Model Training & Architecture Design & \texttt{autoresearch} LLM training
codebase & Final loss ($\downarrow$); test averages two seeds \\
Harness Engineering & Terminal-Bench 2.0 & Official terminal-agent codebase &
Pass rate ($\uparrow$); 36 dev / 53 test tasks \\
Harness Engineering & BrowseComp & Minimal ReAct-style search harness &
Accuracy ($\uparrow$); 50 dev / 300 test questions \\
Data Synthesis & Search-Agent Data Synthesis & Hand-designed search-data
pipeline & Mean pass gap ($\uparrow$); 50 dev / 100 test seeds \\
Data Synthesis & Math-Reasoning Data Synthesis & Hand-designed math-data
pipeline & Mean pass gap ($\uparrow$); 50 dev / 96 test problems \\
\bottomrule
\end{tabularx}
\end{table}

\subsection{Experimental Setup}
\label{sec:exp-setup}

\paragraph{Benchmarks.}
Our evaluation uses two complementary types of benchmarks. The first is the AO
Task Suite in Section~\ref{sec:ao-task-suite}, which consists of real research
tasks with task-specific materials, objectives, development evaluators, and
held-out test evaluators. The second is MLE-Bench Lite, a long-horizon machine
learning engineering benchmark derived from MLE-bench~\citep{mlebench}, which
allows comparison against established benchmark systems under the official task
setup and reporting protocol. 

\paragraph{Baselines.}
For the real research tasks, we compare against two strong coding-agent
baselines: Codex~\citep{codex} using GPT-5.5 and Claude Code~\citep{claudecode} using Claude Opus 4.6. Each
baseline receives the same initial material, objective, evaluator, and resource
budget as \arbor{}, and is allowed to inspect files, edit code, run experiments,
and iterate until the budget is exhausted. For MLE-Bench Lite, we compare
against reported benchmark systems, including AIDE~\citep{aide},
ML-Master~\citep{mlmaster} and ML-Master 2.0~\citep{mlmaster2},
AIRA-dojo~\citep{airadojo}, InternAgent~\citep{internagent},
R\&D-Agent~\citep{rdagent}, Famou-Agent 2.0~\citep{famouagent},
MARS~\citep{mars}, Leeroo~\citep{kapso},
AIBuildAI~\citep{aibuildai}, LoongFlow~\citep{loongflow}, and
AI-Scientist-style systems~\citep{aiscientist,aiscientist_long}. The baseline numbers in Table~\ref{tab:mle-lite} are adopted from the
official MLE-Bench leaderboard and the AI-Scientist paper~\cite{aiscientist_long}.

\paragraph{Metrics.}
We report native task metrics in the main results, using the direction indicated
in Table~\ref{tab:ao-tasks}. For cross-task averages and ablations, we also
report a normalized held-out improvement over the initial material after
orienting all metrics so larger is better. For $\Delta$ rows,
percentage-valued metrics use absolute changes; non-percentage metrics such as
steps and loss use the relative improvement below:
\[
  \Delta_{\test}(\Mat^\star)
  = \frac{\tilde{S}_{\test}(\Mat^\star)-\tilde{S}_{\test}(\Mat_0)}
         {|\tilde{S}_{\test}(\Mat_0)|+\epsilon},
\]
where $\tilde{S}$ is the native score for higher-is-better metrics and the
negated native score for lower-is-better metrics. To measure reliability, we run
each stochastic method three times and report Avg@3 with standard deviation
unless otherwise specified. For MLE-Bench Lite, we report the official benchmark
metrics, including valid-submission rate, above-median rate, any-medal rate, and
medal breakdown.

\paragraph{Implementation details.}
Unless otherwise noted, both the coordinator and executors use
Claude Opus 4.6 as the backbone model. All real-research-task runs, including
Codex~\citep{codex}, Claude Code~\citep{claudecode}, and \arbor{}, use a 48-hour wall-clock limit. To keep the
two single-agent baselines running over this long horizon without manual
intervention, we launch Codex and Claude Code through their official
\texttt{/goal} mode, which lets each agent autonomously sustain a long-running
task and avoid mid-trajectory interruptions; \arbor{} is launched through its
own coordinator loop. The default \arbor{} budget is 20 coordinator cycles
with maximum tree depth 2. Executor parallelism is bounded by the available
evaluator resources, and all wall-clock time, token usage, and evaluator calls
are counted when comparing against baselines. The same prompt-level task
description is used across methods; \arbor{} receives no task-specific search
strategy beyond the adapter that runs the evaluator and parses scores. For
MLE-Bench Lite, every task is optimized on a single NVIDIA A100 GPU under the
official benchmark resource budget.

\subsection{Main Results on Real Research Tasks}
\label{sec:exp-main}

\begin{table}[t]
\centering
\small
\setlength{\tabcolsep}{1.8pt}
\renewcommand{\arraystretch}{1.12}
\caption{Main results on real research tasks. Each task reports native
development and held-out test metrics for the initial material, single-agent
baselines, and \arbor{}; the task label shows the native metric direction.
Shaded $\Delta$ rows report relative improvements over the initial material
for Model Training tasks, and absolute changes for all other tasks.}
\label{tab:main-ao}
\begin{tabularx}{\linewidth}{@{}>{\centering\arraybackslash}p{0.14\linewidth}>{\centering\arraybackslash}p{0.18\linewidth}*{8}{>{\centering\arraybackslash}X}@{}}
\toprule
\multirow[c]{2}{*}{Type} & \multirow[c]{2}{*}{Task} & \multicolumn{2}{c}{Initial} & \multicolumn{2}{c}{Codex} & \multicolumn{2}{c}{Claude Code} & \multicolumn{2}{c}{\arbor{} (Ours)} \\
\cmidrule(lr){3-4}\cmidrule(lr){5-6}\cmidrule(lr){7-8}\cmidrule(l){9-10}
& & Dev & Test & Dev & Test & Dev & Test & Dev & Test \\
\midrule
\multirow[c]{4}{*}{\makecell[c]{Model\\Training}} & \makecell[c]{Optimizer\\Design\\(steps $\downarrow$)} & 3325 & 3325 & 3325 & 3325 & 3275 & 3287.5 & \textbf{3225} & \textbf{3237.5} \\
& \deltalabel & \deltacell{--} & \deltacell{--} & \deltacell{+0.00\%} & \deltacell{+0.00\%} & \deltacell{+1.50\%} & \deltacell{+1.13\%} & \deltacell{\textbf{+3.01\%}} & \deltacell{\textbf{+2.63\%}} \\
& \makecell[c]{Architecture\\Design\\(loss $\downarrow$)} & 1.096 & 1.098 & 1.089 & 1.083 & 1.033 & 1.033 & \textbf{1.029} & \textbf{1.028} \\
& \deltalabel & \deltacell{--} & \deltacell{--} & \deltacell{+0.64\%} & \deltacell{+1.37\%} & \deltacell{+5.75\%} & \deltacell{+5.92\%} & \deltacell{\textbf{+6.11\%}} & \deltacell{\textbf{+6.38\%}} \\
\addlinespace[0.35ex]
\multirow[c]{4}{*}{\makecell[c]{Harness\\Engineering}} & \makecell[c]{Terminal-Bench 2.0\\(pass $\uparrow$)} & 58.33 & 69.81 & 63.89 & 73.59 & \textbf{75.00} & 71.70 & 72.22 & \textbf{77.36} \\
& \deltalabel & \deltacell{--} & \deltacell{--} & \deltacell{+5.56} & \deltacell{+3.78} & \deltacell{\textbf{+16.67}} & \deltacell{+1.89} & \deltacell{+13.89} & \deltacell{\textbf{+7.55}} \\
& \makecell[c]{BrowseComp\\(acc. $\uparrow$)} & 52.50 & 45.33 & 57.50 & 50.00 & 55.00 & 53.33 & \textbf{72.50} & \textbf{67.67} \\
& \deltalabel & \deltacell{--} & \deltacell{--} & \deltacell{+5.00} & \deltacell{+4.67} & \deltacell{+2.50} & \deltacell{+8.00} & \deltacell{\textbf{+20.00}} & \deltacell{\textbf{+22.34}} \\
\addlinespace[0.35ex]
\multirow[c]{4}{*}{\makecell[c]{Data\\Synthesis}} & \makecell[c]{Search-Agent\\(gap $\uparrow$)} & 4.00 & 5.00 & 12.00 & 9.00 & 12.00 & 12.00 & \textbf{16.00} & \textbf{18.00} \\
& \deltalabel & \deltacell{--} & \deltacell{--} & \deltacell{+8.00} & \deltacell{+4.00} & \deltacell{+8.00} & \deltacell{+7.00} & \deltacell{\textbf{+12.00}} & \deltacell{\textbf{+13.00}} \\
& \makecell[c]{Math-Reasoning\\(gap $\uparrow$)} & 2.00 & 1.04 & 6.00 & 6.25 & 8.00 & 8.33 & \textbf{24.00} & \textbf{20.83} \\
& \deltalabel & \deltacell{--} & \deltacell{--} & \deltacell{+4.00} & \deltacell{+5.21} & \deltacell{+6.00} & \deltacell{+7.29} & \deltacell{\textbf{+22.00}} & \deltacell{\textbf{+19.79}} \\
\bottomrule
\end{tabularx}
\end{table}

Table~\ref{tab:main-ao} compares Arbor with Codex and Claude Code on six real research tasks. We focus on two observations.

\paragraph{Arbor gives stronger and more general held-out gains.}
Arbor obtains the best held-out result on all six tasks, covering three different types of research artifacts: training algorithms, agent harnesses, and data-generation pipelines. The same controller and hypothesis-tree depth are used across these tasks; only the initial material and evaluator are changed. This suggests that the improvement comes from the search procedure itself rather than from task-specific tuning.
The gains are also larger than those of single-trajectory coding agents. On BrowseComp, Arbor improves held-out accuracy from $45.33$ to $67.67$, while Codex and Claude Code reach $50.00$ and $53.33$. On Math-Reasoning Data Synthesis, Arbor improves the held-out pass-gap by $19.79$ points, compared with $5.21$ and $7.29$ points for Codex and Claude Code. The baselines can still make progress, but their gains are smaller and less stable across task types. This supports our main hypothesis: in AO, the bottleneck is not only local code editing, but organizing many trials into a coherent exploration process. The cost results in Section~\ref{sec:exp-cost} further show that Arbor achieves these gains without relying on substantially larger token budgets.

\paragraph{The dev/test split exposes overfitting during autonomous search.}
Development feedback is useful for guiding exploration, but it is not a reliable admission criterion. Because the agent repeatedly optimizes against $E_{\mathrm{dev}}$, it can overfit to the development split or exploit evaluator-specific patterns. This is especially visible on Terminal-Bench: Claude Code achieves the highest development score ($75.00$), but its held-out score drops to $71.70$; Arbor has a lower development score ($72.22$), but reaches the best held-out score ($77.36$).
This gap motivates the held-out merge gate in Arbor. We use $E_{\mathrm{dev}}$ to guide hypothesis search, but promote a candidate artifact only when it improves $E_{\mathrm{test}}$. This separates exploratory feedback from verified progress. It also makes dev/test disagreement informative: a high-dev, low-test candidate is treated not as a success, but as evidence that the current direction may be exploiting the feedback signal rather than producing a transferable improvement.

\begin{table}[t]
\centering
\scriptsize
\setlength{\tabcolsep}{3.0pt}
\renewcommand{\arraystretch}{1.08}
\caption{MLE-Bench Lite results under the official evaluation protocol. All
entries are percentages.}
\label{tab:mle-lite}
\resizebox{0.93\linewidth}{!}{%
\begin{tabular}{llcccccc}
\toprule
Method & Model & \makecell{Valid sub.} & \makecell{Above\ median} & Bronze & Silver & Gold & \makecell{Any\ medal} \\
\midrule
InternAgent & DeepSeek-R1 & 100.00 & 78.79 & 10.61 & 16.67 & 34.85 & 62.12 \\
ML-Master & DeepSeek-R1 & 100.00 & 74.24 & 4.55 & 13.64 & 30.30 & 48.48 \\
AIRA-dojo & o3 & 100.00 & 70.45 & 7.95 & 12.73 & 34.32 & 55.00 \\
ML-Master 2.0 & DeepSeekV3.2-Spe & 100.00 & 84.85 & 13.64 & 31.82 & 30.30 & 75.76 \\
R\&D-Agent & GPT-5 & 77.27 & 74.24 & 12.12 & 22.73 & 33.33 & 68.18 \\
Famou-Agent 2.0 & Gemini-2.5-Pro & 100.00 & 86.36 & 15.15 & 19.70 & 40.91 & 75.76 \\
MARS & Gemini-3-Pro & 100.00 & 89.39 & 6.06 & 15.15 & 53.03 & 74.24 \\
Leeroo & Gemini-3-Pro & 68.18 & 68.18 & 18.18 & 19.70 & 30.30 & 68.18 \\
AIBuildAI & Claude-Opus-4.6 & 100.00 & 81.82 & 13.64 & 25.76 & 37.88 & 77.27 \\
AIDE & Gemini-3-Flash & 77.27 & 54.55 & 4.55 & 9.09 & 31.82 & 45.45 \\
LoongFlow & Gemini-3-Flash & 77.27 & 77.27 & 12.12 & 25.76 & 39.39 & 77.27 \\
Codex & GPT-5.5 (xhigh) & 100.00 & 81.82 & 1.52 & 19.70 & 46.97 & 68.18 \\
AI-Scientist & Gemini-3-Flash & 100.00 & 86.36 & 18.18 & 31.82 & 31.82 & \underline{81.82} \\
\midrule
\rowcolor{selfevolagent_lighter!55}
\arbor{} & Gemini-3-Flash & 100.00 & 86.36 & 13.64 & 27.27 & 40.90 & \underline{81.82} \\
\rowcolor{selfevolagent_lighter!55}
\arbor{} & GPT-5.5 & 100.00 & 95.45 & 0.00 & 9.09 & 77.27 & \textbf{86.36} \\
\bottomrule
\end{tabular}
}
\end{table}

\subsection{Results on MLE-Bench Lite}
\label{sec:exp-mle}

We also evaluate \arbor{} on MLE-Bench Lite under the official protocol.
Unlike our AO task suite, this benchmark fixes the competition-style ML tasks,
scoring rules, and medal thresholds, so it tests whether the same controller can
turn repeated experiments into stronger runnable submissions. \arbor{} uses the
same controller as before, adding only an adapter for workspace setup and
submission formatting. Table~\ref{tab:mle-lite} reports the results.

With a matched Gemini-3-Flash backbone, \arbor{} reaches 100\% valid
submissions, 86.36\% above-median rate, and 81.82\% any-medal rate, tying the
best same-backbone any-medal result while obtaining a higher gold rate than
AI-Scientist and LoongFlow. Replacing the backbone with GPT-5.5, without
changing the controller, depth, scheduler, or adapter, further raises any-medal
to 86.36\% and gold to 77.27\%, the highest values in
Table~\ref{tab:mle-lite}. These results suggest that the hypothesis-tree organization transfers beyond
our constructed AO tasks to established long-horizon ML engineering benchmarks.

\subsection{Backbone Generality}
\label{sec:exp-generality}

\begin{figure}[t]
    \centering
  \includegraphics[width=\linewidth]{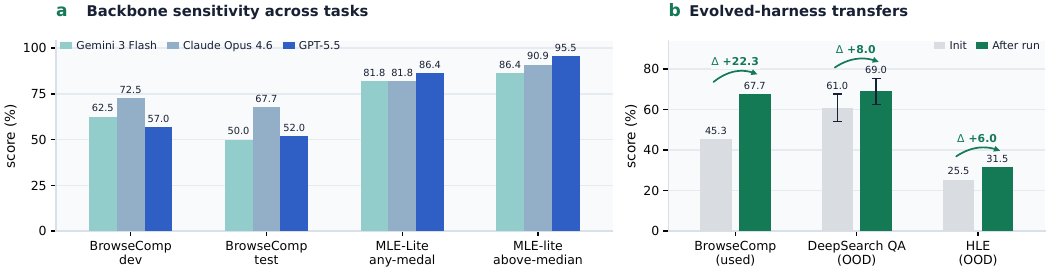}
  \caption{Backbone generality and cross-task transfer. \textbf{(a)}
  \arbor{} is rerun with different backbone models while keeping the
  controller, evaluator budget, and task adapters fixed. \textbf{(b)}
  A BrowseComp-evolved search harness is frozen and evaluated on held-out
  search-agent tasks without further task-specific optimization.}
    \label{fig:section45-46-results}
\end{figure}

We next test whether \arbor{}'s gains are tied to a particular backbone model. We repeat representative
runs with different backbones. As shown in Figure~\ref{fig:section45-46-results}(a), \arbor{} is not tied to a single frontier model: even with Gemini-3-Flash, a lighter backbone than the Claude and GPT variants, the same controller still improves both browsecomp and MLE-Bench Lite. This suggests that \htr{} provides a
model-agnostic structure for exploration and memory rather than depending on a
specific model.

We also notice that backbone effects are task-dependent. Although
\htr{} provides a common structure for exploration and memory, final
performance is mediated by the compatibility between a model's capabilities and
the task requirements. Claude Opus 4.6 performs best on BrowseComp, where
improving a search harness relies heavily on broad reasoning and error
diagnosis. In contrast, GPT-5.5 performs best on MLE-Bench Lite, where gains are
more closely tied to ML-engineering knowledge, including data processing,
training recipes, and leaderboard-oriented optimization. Thus, \arbor{} is
model-agnostic at the framework level, but its empirical ceiling depends on
task--backbone fit.

\subsection{Cross-Task Idea Transfer}
\label{sec:exp-transfer}

A stronger test of generality is whether an optimized artifact transfers beyond the benchmark used for search. This is important for auto-research and auto-harness systems: an agent may improve on a source evaluator by exploiting benchmark-specific patterns rather than discovering generally useful design changes.

We therefore evaluate transfer in the harness-engineering setting. \arbor{} is first run on BrowseComp, using only BrowseComp development feedback to propose, implement, and merge search-harness changes. After the run, we freeze the resulting harness and evaluate it directly on two unseen search-agent tasks, HLE and DeepSearchQA, without further task-specific optimization.

Figure~\ref{fig:section45-46-results}(b) shows that the learned harness transfers. The optimized harness improves BrowseComp held-out accuracy from 45.33\% to 67.67\%. More importantly, the same frozen codebase also improves HLE from 25.50\% to 31.50\% and DeepSearchQA from $61.00\pm6.76\%$ to $69.00\pm6.41\%$. Since these two tasks are never used during BrowseComp optimization, the gains indicate that Arbor can discover harness-level changes that survive a shift in task distribution, rather than only fitting the source benchmark.

\begin{table}[!t]
\centering
\begin{minipage}[t]{0.53\linewidth}
\vspace{0pt}
\centering
\scriptsize
\setlength{\tabcolsep}{1.3pt}
\renewcommand{\arraystretch}{1.08}
\caption{Component ablations on MLE-Bench Lite (Claude Opus 4.6 backbone).
Entries are percentages.}
\label{tab:mle-ablation}
\begin{tabular*}{\linewidth}{@{\extracolsep{\fill}}lcccccc@{}}
\toprule
Variant & \makecell{Valid\\sub.} & \makecell{Above\\median} & Bronze & Silver & Gold & \makecell{Any\\medal} \\
\midrule
Full \arbor{} & 100.00 & 90.91 & 4.55 & 27.27 & 50.00 & \textbf{81.82} \\
w/o tree & 100.00 & 72.72 & 9.09 & 22.73 & 31.82 & 63.64 \\
w/o insight feedback & 100.00 & 77.27 & 4.55 & 13.64 & 36.36 & 54.54 \\
\bottomrule
\end{tabular*}
\end{minipage}\hfill
\begin{minipage}[t]{0.43\linewidth}
\vspace{0pt}
\centering
\scriptsize
\setlength{\tabcolsep}{1.2pt}
\renewcommand{\arraystretch}{1.08}
\caption{\arbor{}'s node statistics. Dev+ means nodes that improve over the baseline on the dev set.}
\label{tab:search-trace}
\begin{tabular*}{\linewidth}{@{\extracolsep{\fill}}lcccccc@{}}
\toprule
\makecell[l]{Node\\num.} & \makecell[c]{Opt.\\Design} & \makecell[c]{Arch.\\Design} & \makecell[c]{Terminal\\Bench} & \makecell[c]{Browse\\Comp} & \makecell[c]{Search\\Agent} & \makecell[c]{Math\\Reason.} \\
\midrule
All & 26 & 150 & 17 & 26 & 15 & 15 \\
Dev+ & 13 & 15 & 7 & 10 & 10 & 6 \\
Merged & 2 & 9 & 3 & 3 & 4 & 4 \\
\bottomrule
\end{tabular*}
\end{minipage}
\end{table}

\subsection{Ablations}
\label{sec:exp-ablation}

We ablate the two components most central to HTR on MLE-Bench Lite: the hierarchical hypothesis tree and insight feedback. The \textit{w/o tree} variant reduces search to a flat experiment queue, with all experiments attached directly to the root. The \textit{w/o insight feedback} variant keeps the tree structure but disables upward propagation of distilled lessons. Both variants use the same tool access, workspace budget, evaluation protocol, and Claude Opus 4.6 backbone as the full system.

\paragraph{HTR improves refinement rather than executability.}
Table~\ref{tab:mle-ablation} shows that all variants obtain 100\% valid submissions, indicating that the ablation gap is not caused by basic execution failure. The difference instead appears in outcome quality. Full Arbor reaches 81.82\% Any Medal, compared with 63.64\% for \textit{w/o tree} and 54.54\% for \textit{w/o insight feedback}. The same pattern appears in stronger categories such as Above Median, Silver, and Gold. This suggests that HTR mainly improves later-stage research refinement: once a runnable solution exists, the tree helps the agent decide which directions to extend, revise, or abandon.

\paragraph{The tree is useful only when evidence can accumulate over it.}
Removing insight feedback while keeping the tree causes a larger drop than removing the tree entirely. This result suggests that hierarchy alone is not sufficient. A tree without propagated lessons can still organize experiments syntactically, but it does not provide the semantic memory needed for later decisions. In contrast, full Arbor uses the tree as a substrate for accumulating evidence: leaf-level results are abstracted into direction-level lessons, which then constrain future ideation and selection.

\paragraph{Tree structure and insight feedback are complementary.}
The full system outperforms both ablations, indicating that the two components address different parts of the search problem. The tree defines where competing hypotheses are stored and compared, while insight feedback determines what reusable information is carried forward. Their combination allows Arbor to convert local experimental outcomes into persistent constraints on future search, rather than treating each experiment as an isolated trial.

\subsection{Token Consumption and Search Cost}
\label{sec:exp-cost}

\begin{figure}[!t]
\centering
\includegraphics[width=0.88\linewidth]{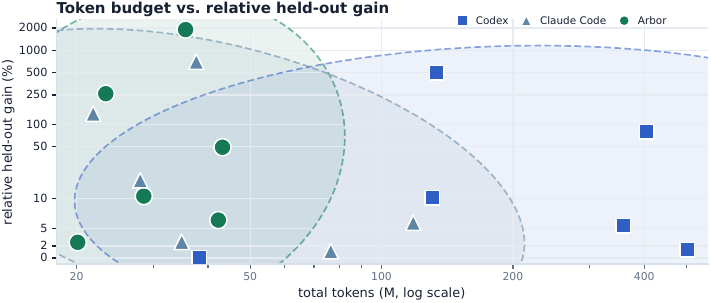}
\caption{Token budget and relative held-out gain across completed AO cost logs. Token totals sum input
and output tokens; for \arbor{}, the total further sums coordinator and executor
usage. The y-axis reports percent improvement over each task's initial held-out
score.}
\label{fig:token-cost-gain}
\end{figure}

We further examine whether \arbor{}'s gains mainly come from increased model budget. Figure~\ref{fig:token-cost-gain} reports total token consumption and relative held-out gain, while Table~\ref{tab:search-trace} summarizes the corresponding tree traces.

\paragraph{Structured search rather than larger sampling.}
Across the six completed cost logs, \arbor{} uses 20.12M--43.19M tokens, a comparable scale to the single-trajectory baselines. Within this budget, \arbor{} achieves larger held-out gains on most tasks. This suggests that the improvement is not simply due to spending substantially more tokens, but to how the budget is organized: tokens are used to maintain competing hypotheses, run isolated executions, compare evidence, and update the search tree.

\paragraph{Dev improvements are filtered by held-out admission.}
Table~\ref{tab:search-trace} also shows that many nodes improve the development score, but only a smaller subset are merged. This gap is expected. A dev-improving node may still be worse than the current best artifact, or may overfit the development evaluator and fail to transfer to the held-out test. The merge gate therefore prevents local development gains from being mistaken for artifact-level progress. In this sense, the tree trace records broad exploration, while the held-out gate admits only verified improvements into the final artifact.

\begin{figure}[t]
\centering
\includegraphics[width=\linewidth]{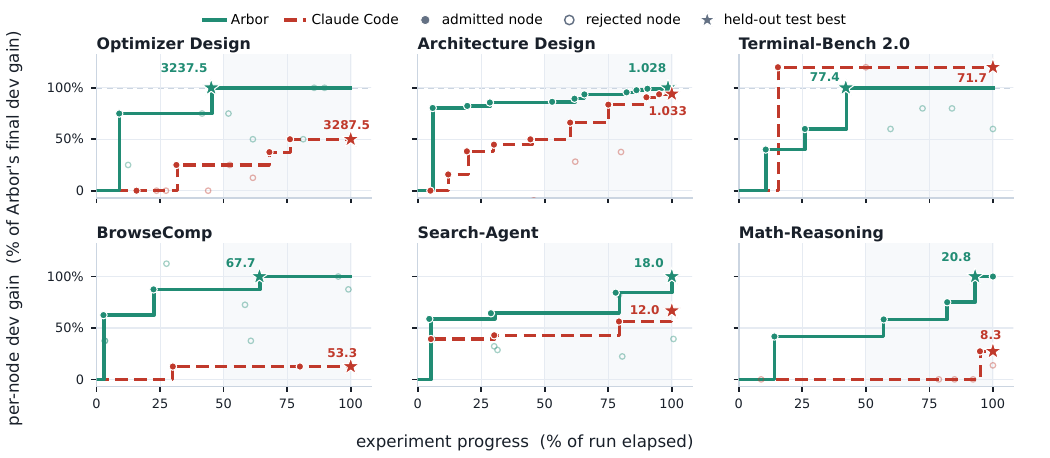}
\caption{\textbf{Exploration efficiency on the six AO tasks} (one panel per
task). Curves show the best-so-far development gain over the run, normalized to
\arbor{}'s final gain, so \arbor{} (solid) ends at $100\%$ and the Claude Code
baseline (dashed) at its own relative ceiling. Stars mark each method's
held-out test maximum, annotated with the test score from
Table~\ref{tab:main-ao}. }
\label{fig:discussion-insights}
\end{figure}

\section{Discussion}
\label{sec:discussion}

We analyze \arbor{}'s internal research traces to understand how autonomous research progresses once the agent starts running experiments. We focus on three questions: how hypotheses change over time (Section~\ref{sec:discussion-refinement}), when useful improvements appear (Section~\ref{sec:discussion-timing}), and what kinds of ideas the Hypothesis Tree produces (Section~\ref{sec:discussion-ideas}).
 
\subsection{Hypothesis Refinement Analysis}
\label{sec:discussion-refinement}
 
\begin{figure}[t]
\centering
\includegraphics[width=\linewidth]{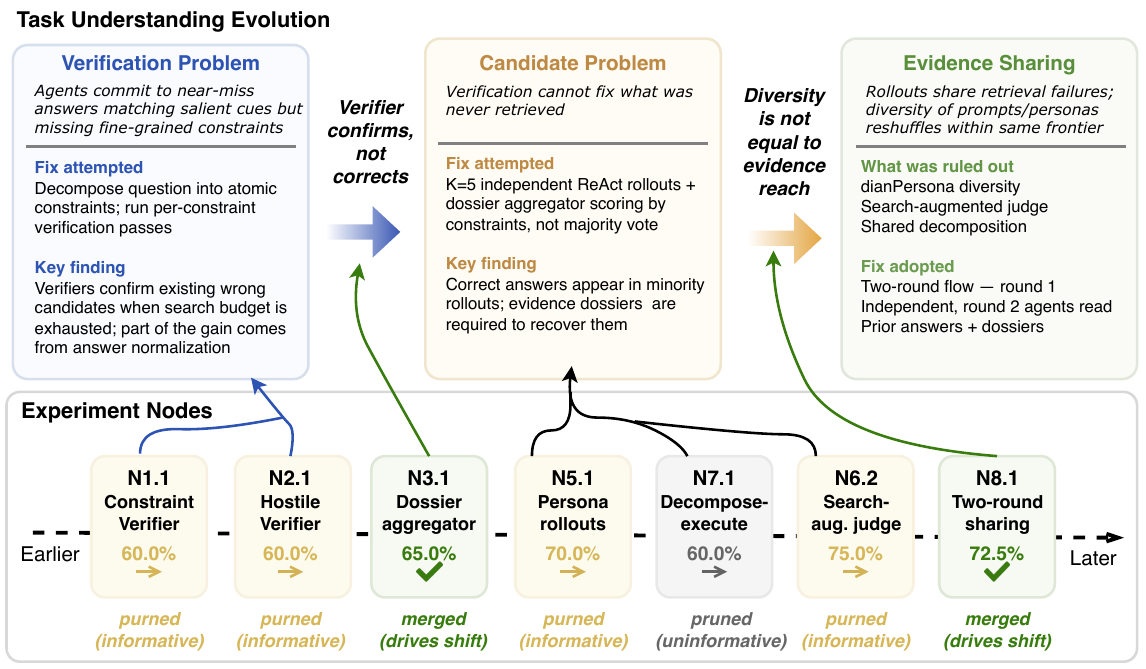}
\caption{Evolution of task understanding across the BrowseComp run.
Each upper-tier box states the current problem framing, the fix attempted, and the mechanistic finding that drove the next shift.
The lower tier shows the experimental nodes behind each transition.}
\label{fig:bc-insight-evolution}
\end{figure}
 
We analyze the BrowseComp hypothesis tree, reporting the main hypothesis shifts, the nodes that triggered them, and the final design selected by the merge gate. Figure~\ref{fig:bc-insight-evolution} traces all three contractions of task understanding alongside the experimental nodes that drove each transition. We find that:
 
\textbf{Early nodes test whether a broad mechanism holds.}
The run begins from a coarse hypothesis and uses the first experiments to confirm or reject it. In BrowseComp, the initial hypothesis is that search agents produce near-miss answers by matching salient cues while missing fine-grained constraints; constraint-decomposed verification and hostile-contradiction checking both improve development accuracy, confirming that fine-grained answer checking is a valid source of gain.
 
\textbf{Later nodes localize the bottleneck by probing the mechanism's boundary.}
Once a mechanism is confirmed, the tree tests where it stops working rather than pushing it further. In BrowseComp, the verifier nodes can judge candidates produced by the search process but rarely recover candidates that were never surfaced, and part of the hostile-verifier gain comes from answer normalization rather than reliable evidence discovery. This shifts the main design target from stricter verification to broader evidence coverage.
 
\textbf{Ancestor insights compress these results into the constraints that shape the final design.}
The accumulated positive and negative findings define what the successful design must satisfy. In BrowseComp, the evidence-dossier aggregator preserves candidates and supporting evidence across independent rollouts, recovering correct answers that appear in only a minority of trajectories; follow-up nodes then rule out persona-diverse rollouts (which only rerank within the same retrieval frontier), the search-augmented judge (which overfits development questions), and shared decomposition (which reduces trajectory independence). \arbor{} thus learns that BrowseComp benefits from sharing evidence while keeping search trajectories independent.
 
\paragraph{Takeaway.}
Hypothesis refinement in HTR is a \textbf{deepening of task understanding}: early nodes test broad mechanisms, later nodes identify their limits, and ancestor insights summarize these results into constraints for the next round of proposals. This constraint accumulation is the core process-level benefit over flat trial-and-error.
 
\subsection{Search Efficiency Analysis}
\label{sec:discussion-timing}
 
We analyze when the best candidates appear during a run, using each node's execution time and development gain. We find that:
 
\textbf{Strong candidates often appear after the search state has accumulated constraints.}
Across tasks, \arbor{} frequently reaches its best candidate in the middle or later part of the run. These improvements are supported by earlier nodes that identify useful mechanisms, rule out weak variants, and narrow the design space.
 
\textbf{Later proposals are more targeted than early proposals.}
In BrowseComp, the final evidence-sharing design appears only after several failed or partially successful verifier variants establish that fine-grained checking matters, that candidate coverage is the bottleneck, and that judge-side search and shared decomposition introduce new failure modes. The final proposal is therefore generated from a more informative research state than the initial proposals.
 
\textbf{The tree improves search by changing the proposal distribution over time.}
\arbor{} does not simply allocate budget uniformly over independent attempts: its later nodes are conditioned on accumulated evidence from ancestors and siblings. Successful mechanisms become priors, failed variants become negative constraints, and partial gains become starting points for refined hypotheses.
 
\paragraph{Takeaway.}
\textbf{Earlier experiments persistently reduce the arbitrariness of later search, placing mid-to-late improvements on a higher information baseline.}
The relevant notion of efficiency is whether the same budget produces a less repetitive and more constrained evidence chain, rather than running long enough to stumble onto a result by chance.
 
\subsection{Idea Quality Analysis}
\label{sec:discussion-ideas}
 
\begin{figure}[t]
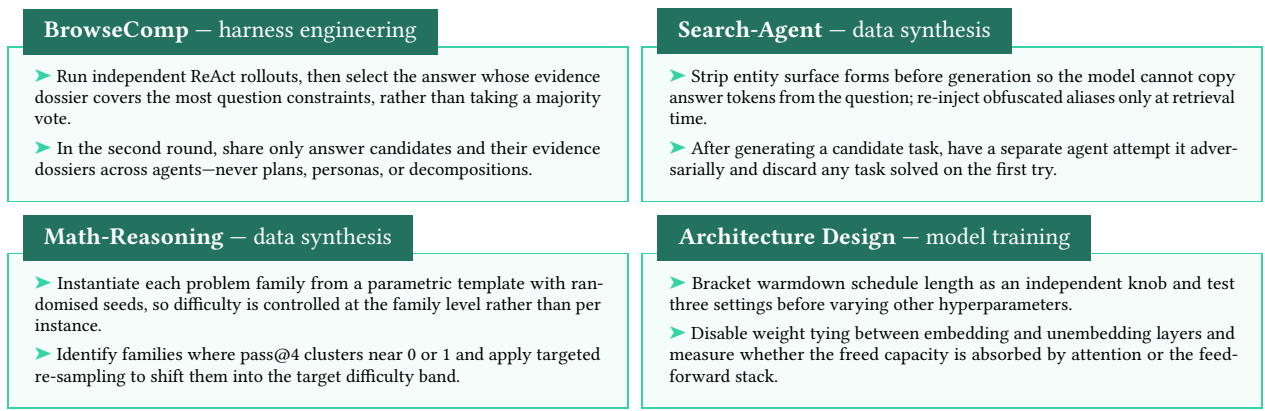

\centering
\newtcolorbox{ideacard}[1]{%
  enhanced, breakable=false, sharp corners=all,
  colback=selfevolagent_lighter!22, colframe=selfevolagent_dark,
  boxrule=0.6pt, left=7pt, right=7pt, top=5pt, bottom=5pt,
  fonttitle=\bfseries\small\color{white},
  coltitle=white, colbacktitle=deltaaccent,
  attach boxed title to top left={xshift=6pt, yshift=-3pt},
  boxed title style={sharp corners=all, boxrule=0pt, left=5pt, right=5pt, top=1.5pt, bottom=1.5pt},
  title={#1}, fontupper=\scriptsize, before upper=\setlength{\parskip}{3pt},
}
\newcommand{\idea}[1]{\textcolor{selfevolagent_dark}{\ding{228}}~#1\par}
\begin{tcbraster}[raster columns=2, raster column skip=5pt, raster row skip=5pt,
                  raster equal height=rows, raster width=\linewidth]
  \begin{ideacard}{BrowseComp \textnormal{\textemdash{} harness engineering}}
    \idea{Run independent ReAct rollouts, then select the answer whose evidence dossier covers the most question constraints, rather than taking a majority vote.}
    \idea{In the second round, share only answer candidates and their evidence dossiers across agents---never plans, personas, or decompositions.}
  \end{ideacard}
  \begin{ideacard}{Search-Agent \textnormal{\textemdash{} data synthesis}}
    \idea{Strip entity surface forms before generation so the model cannot copy answer tokens from the question; re-inject obfuscated aliases only at retrieval time.}
    \idea{After generating a candidate task, have a separate agent attempt it adversarially and discard any task solved on the first try.}
  \end{ideacard}
  \begin{ideacard}{Math-Reasoning \textnormal{\textemdash{} data synthesis}}
    \idea{Instantiate each problem family from a parametric template with randomised seeds, so difficulty is controlled at the family level rather than per instance.}
    \idea{Identify families where pass@4 clusters near 0 or 1 and apply targeted re-sampling to shift them into the target difficulty band.}
  \end{ideacard}
  \begin{ideacard}{Architecture Design \textnormal{\textemdash{} model training}}
    \idea{Bracket warmdown schedule length as an independent knob and test three settings before varying other hyperparameters.}
    \idea{Disable weight tying between embedding and unembedding layers and measure whether the freed capacity is absorbed by attention or the feed-forward stack.}
  \end{ideacard}
\end{tcbraster}
\caption{Representative ideas generated by \arbor{} across tasks. Each idea is procedural rather than instance-specific, and each was proposed only after preceding nodes had ruled out broader directions.}
\label{fig:representative-ideas}
\end{figure}
 
We analyze representative ideas generated across model training, harness engineering, and data synthesis tasks, classifying each by its granularity, implementation target, and relation to previous evidence. Figure~\ref{fig:representative-ideas} shows representative examples. We find that:
 
\textbf{Most useful ideas are local and executable.}
In model training, ideas usually modify a specific optimizer component, training recipe, or architecture choice. In harness engineering, they change concrete parts of the agent loop, such as retrieval, aggregation, verification, or context management. In data synthesis, they refine generation, filtering, difficulty calibration, or verification modules. This locality makes each idea easy to implement, evaluate, and attribute to a tree node.
 
\textbf{Useful ideas are often evidence-conditioned.}
Many successful proposals directly respond to earlier observations: the BrowseComp evidence-dossier design follows from the failure mode of verifier-only approaches, and similar patterns appear in data synthesis, where later nodes repair specific weaknesses in difficulty calibration or answer verification. HTR therefore helps convert local failures into new design constraints, and ensures that ``half-right'' results become the starting point for a more precise hypothesis rather than a reason to abandon the direction.
 
\textbf{High-level problem formulation remains important.}
\arbor{} is strongest when the objective can be improved through a sequence of concrete refinements, and less reliable when progress requires a new high-level formulation weakly connected to the existing tree. The Architecture Design task ultimately acknowledged that single-knob tuning had reached diminishing returns and a larger algorithmic move was needed, but identifying that move still depended on prior judgment rather than anything the tree could automatically generate. This highlights the role of human-provided task design: the initial artifact, evaluator, metric, and search interface shape the kinds of ideas the agent can discover.
 
\paragraph{Takeaway.}
As the tree grows, what has been ruled out, validated, and found to have boundary conditions all become priors constraining the next round of proposals.
\arbor{}'s ideas are therefore not isolated guesses but local advances relative to a known task understanding. Together with the refinement and timing results above, this paints the complete picture of HTR as a process mechanism that makes autonomous research cumulative: not more attempts, but less repetitive and more memory-aware search.
 
\FloatBarrier

\section{Conclusion}\label{sec:conclusion}

We presented \arbor{} as a framework for Autonomous Optimization, where a
research agent must improve a real artifact through long-horizon experimental
feedback rather than execute a single predefined trajectory. The core idea is
to make the research state persistent and operational: \arbor{} represents
competing hypotheses, artifact versions, evaluation results, failure
attributions, and reusable insights in a durable hypothesis tree. A coordinator
uses this tree to manage strategic search, while short-lived executors ground
individual hypotheses in isolated worktrees and return structured evidence.
Together with insight propagation and a held-out admission gate, this design
turns trial and error into an auditable process of branching, falsification,
and evidence-constrained improvement.

Across the AO settings studied here, this organization provides consistent
evidence of value. On six real-research tasks spanning model training, harness
engineering, and data synthesis, \arbor{} achieves the strongest held-out
results among the compared methods; on MLE-Bench Lite, the same controller
transfers to an established long-horizon ML-engineering benchmark. The transfer
study shows that a BrowseComp-optimized harness can improve unseen search-agent
tasks, and the ablations indicate that the hypothesis tree and insight feedback
are most useful when they operate together. These results support the view that
persistent hypothesis management is a useful abstraction for autonomous
research, while the limitations of the current task suite, scalar objectives,
model capabilities, and search cost leave substantial room for broader and more
rigorous future evaluations.

\FloatBarrier

\appendix

\newpage
\section*{Appendix}
\label{app:contents}
\providecommand{\ours}{\arbor}
\startcontents[sections]
\setupappendixtoc
{\hypersetup{linkcolor=AppendixTOCSection}
\printcontents[sections]{l}{1}{\setcounter{tocdepth}{3}}}

\section{Limitations and Future Work}
\label{app:limitations-future}

Although the empirical results demonstrate the promise of \arbor{} for
autonomous research, this study has several limitations. We discuss these
limitations below and outline the corresponding directions for future work.

\paragraph{Evaluation scope.}
Our experiments are an initial probe of autonomous research rather than a
complete benchmark for scientific discovery. The current AO task suite covers
model training, harness engineering, and data synthesis, but it does not yet
span the full diversity of research problems. Within AI, future tasks should
include settings such as low-level kernel optimization, pretraining data-mixture
design, and more open-ended system design. Beyond AI, domains such as biology,
mathematics, and physics require benchmarks where valuable hypotheses are harder
to specify and validate. A broader suite should therefore evaluate not only
metric improvement, but also whether the generated ideas are scientifically
meaningful, reproducible, and transferable.

\paragraph{Objective design.}
The present AO interface mainly optimizes a fixed scalar objective defined by a
task-specific evaluator. This is useful for controlled experiments, but it is a
simplification of real research. Scientific objectives are often multi-dimensional:
performance, resource use, robustness, interpretability, novelty, and safety may
all matter, and improving one can hurt another. Future AO systems should support
multi-objective search, explicit constraints, Pareto-style comparison, and
adaptive scheduling between competing criteria. This would also reduce the risk
that an agent overfits to a narrow benchmark metric while missing the broader
research goal.

\paragraph{Idea generation.}
We observe that agents can read evaluation feedback carefully and propose useful
local refinements, but their research ability remains far from that of expert
human researchers. In difficult tasks, they may fail to identify a genuinely new
mechanism, abandon a promising direction after early failures, or reverse-engineer
solutions from observed scores instead of reasoning from first principles. A more
fine-grained study of agent idea formation is therefore needed. Promising
directions include better uncertainty tracking, explicit reuse of negative
evidence, mechanisms for revisiting suspended branches, and training or prompting
methods that encourage causal and first-principles hypotheses rather than only
result-driven fixes.

\paragraph{Cost and infrastructure.}
Long-horizon autonomous research is limited not only by idea quality, but also
by systems engineering. In our runs, performance and efficiency depend on details
such as prompt caching, evaluator scheduling, isolated environment startup,
parallel worktree execution, and the reliability of inter-agent coordination.
Large numbers of model calls, evaluator calls, and artifact handoffs can make a
successful search expensive even when each individual step is simple. Future work
should develop cost-aware tree policies, adaptive evaluator allocation, stronger
caching and checkpointing, and more robust execution infrastructure so that AO
systems can scale without turning search breadth into uncontrolled compute cost.

\paragraph{Model capability.}
Finally, \arbor{} inherits the strengths and weaknesses of the underlying LLMs.
Current models are often capable of coding, summarizing results, and making
plausible local hypotheses, but they can still struggle with deep domain
knowledge, long chains of causal reasoning, and genuinely creative problem
reformulation. Stronger foundation models will likely improve AO directly, but
model scaling alone may not be sufficient. Future systems should combine LLMs
with domain knowledge bases, specialized tools, simulators, formal checkers, and
training signals targeted at scientific hypothesis generation. In this sense,
\arbor{} provides a structure for accumulating and testing ideas, while the
quality of those ideas remains an important frontier.

\section{Details of \texorpdfstring{\ours{}}{Arbor}}
\label{app:arbor-details}

\begin{figure}[!t]
  \centering
  \includegraphics[width=\linewidth]{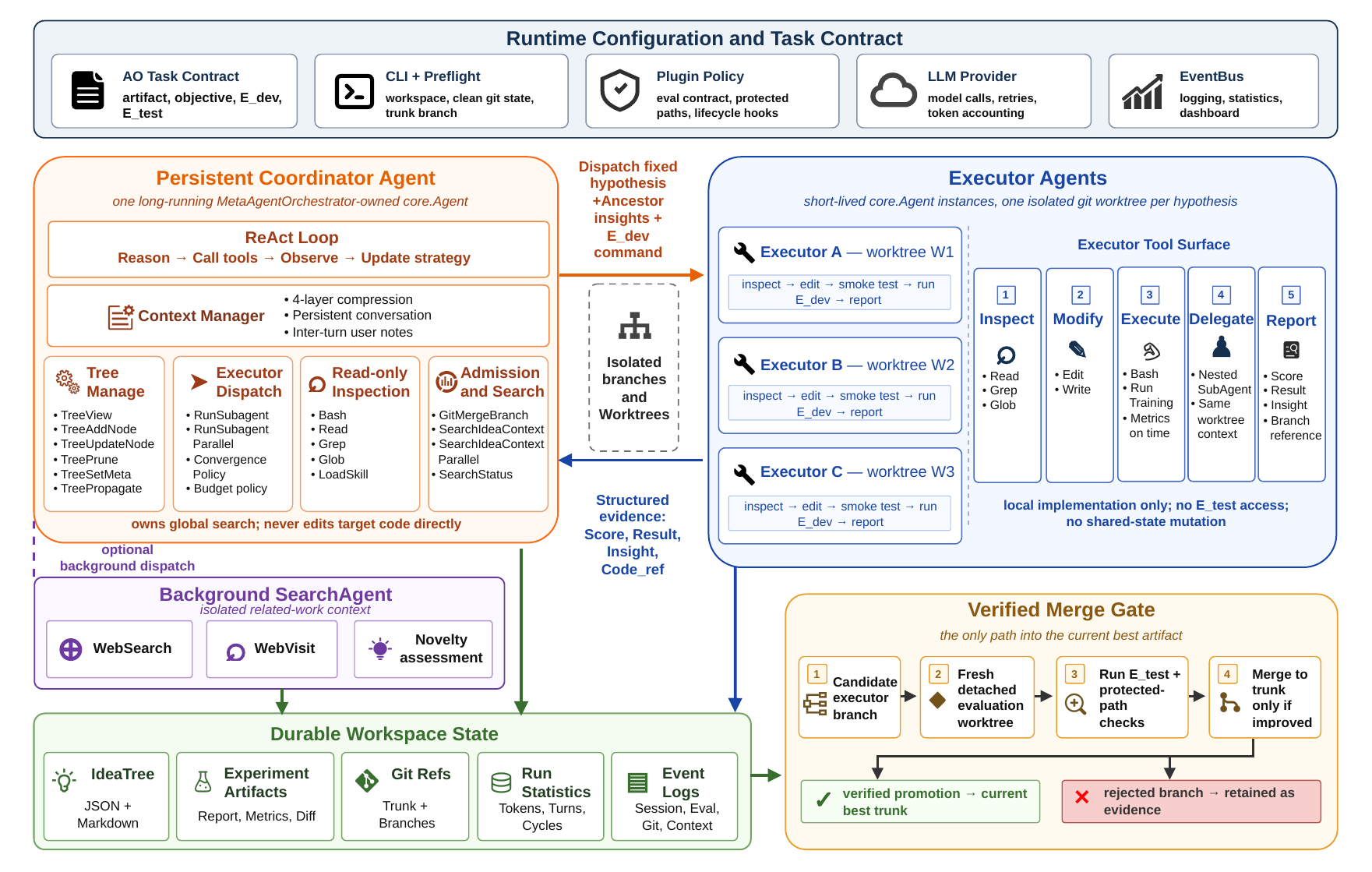}
  \caption{Agent-level implementation details of \ours{}.}
  \label{fig:appendix-overall-framework}
\end{figure}

Figure~\ref{fig:appendix-overall-framework} summarizes the implementation-level
agent internals used by \ours{}. 

\subsection{Prompts}
\label{app:prompts}

\subsubsection{Coordinator Prompt}
\label{app:meta-agent-prompt}

\begin{promptbox}[title=\textbf{Coordinator Prompt}]
\noindent\textbf{Role.}~You are the persistent coordinator in \arbor{}.
Your job is to maintain the hypothesis tree as the shared research state,
choose which hypotheses deserve execution, and convert experimental feedback
into reusable evidence. You do not edit the target artifact directly. All
implementation work is delegated to short-lived Executors, while you own the
tree, the research frontier, and the merge/prune/stop decisions.

\smallskip
\noindent\textbf{Runtime and durable state.}~You run in a single persistent
ReAct loop. When the conversation approaches the context limit, older turns may
be compressed; the on-disk hypothesis tree is therefore the source of truth.
Every controlled mutation is saved as JSON for tools and regenerated as
Markdown for human inspection. Operator messages prefixed with
\texttt{[user note]} are instructions or questions that must be handled at the
next safe point before launching new Executors or merging branches.

\smallskip
\noindent\textbf{Research contract.}~At initialization, inspect the target
codebase and record the AO contract via \texttt{TreeSetMeta}: the objective,
metric direction, development evaluator $\Eval_\dev$, held-out evaluator
$\Eval_\test$, dataset paths, baseline score, and evaluation commands. Commands
must use the template variables \texttt{\{cwd\}} and \texttt{\{node\_id\}} so
that the same contract can be injected safely into isolated executor worktrees.

\smallskip
\noindent\textbf{Hypothesis tree.}~The root node represents the initial
artifact. Depth-1 children are broad research directions; deeper nodes are more
specific refinements, alternatives, or corrections under their parent. Each
node stores a lifecycle status (\texttt{pending}, \texttt{running},
\texttt{done}, \texttt{merged}, or \texttt{pruned}), a dev score, a factual
result, a structured insight, and a git branch reference \texttt{code\_ref}.
The tree is not a transcript of tool calls: it is the compact record of what
was tried, what happened, why it happened, and how that evidence should shape
future hypotheses.

\smallskip
\noindent\textbf{Coordinator loop.}~Repeat the following cycle until the
budget is exhausted, no promising frontier remains, or the objective has been
satisfied.
\begin{enumerate}[label=\textbf{\arabic*.},leftmargin=1.6em,itemsep=2pt,topsep=2pt]
  \item \textbf{Observe.} Read the current tree with \texttt{TreeView}; inspect
  the current best artifact, recent executor reports, and experiment logs when
  needed. Reconstruct the live research frontier from the persisted tree rather
  than from memory alone.
  \item \textbf{Constrain.} Use \texttt{TreeView(format=``constraints'')} to
  load tree shape, root insight, pruned lessons, and validated findings. Treat
  pruned lessons as negative constraints and validated findings as assumptions
  to build on.
  \item \textbf{Ideate.} Select a parent node and propose a small set of
  executable child hypotheses. Each candidate added with \texttt{TreeAddNode}
  must state its mechanism, hypothesis, observable success condition, and known
  conflicts or risks. Do not add candidates that merely repeat a pruned failure
  mode.
  \item \textbf{Select.} Choose pending leaves for the next batch using
  accumulated evidence, expected impact, implementation cost, and diversity
  across active directions. Prefer hypotheses that can teach the tree something
  useful even when they fail.
  \item \textbf{Execute.} Dispatch selected leaves with \texttt{RunSubagent} or
  \texttt{RunSubagentParallel}. Each Executor receives one fixed hypothesis,
  the research contract, and ancestor insights, then works in an isolated git
  worktree branched from the current best artifact.
  \item \textbf{Update.} After execution, write back the dev score, factual
  result, distilled insight, and \texttt{code\_ref}. Propagate insights upward
  so leaf-level outcomes become direction-level lessons and eventually update
  the global prior at the root.
  \item \textbf{Decide.} Continue a direction, prune it with \texttt{TreePrune}
  when its assumptions are falsified, or promote a candidate through
  \texttt{GitMergeBranch}. A branch may update the current best only after the
  merge gate evaluates it on $\Eval_\test$ in a fresh worktree and confirms an
  improvement over the current best.
\end{enumerate}

\smallskip
\noindent\textbf{Executor boundary.}~Executors materialize hypotheses; they do
not control the global search. They may repair implementation mistakes and rerun
$\Eval_\dev$, but they must not change the assigned hypothesis, inspect sibling
branches as shortcuts, or use the held-out evaluator for routine iteration.
Their reports should be compact enough to become tree evidence: score, result,
insight, and branch reference.

\smallskip
\noindent\textbf{Merge and termination rules.}~Use \texttt{GitMergeBranch}
exclusively for promotion; direct git merges through Bash are prohibited. Before
terminating, run the held-out evaluator on both the final best artifact and the
baseline when required, then record \texttt{test\_trunk\_score} and
\texttt{test\_baseline\_score} via \texttt{TreeSetMeta}. Failed, pruned, and
unmerged branches should remain in the tree as reusable evidence rather than
being erased from the research record.
\end{promptbox}

\subsubsection{Executor Prompt}
\label{app:sub-agent-prompt}

\begin{promptbox}[title=\textbf{Executor Prompt}]
\noindent\textbf{Identity.}~You are a Research Agent that implements research
ideas into codebases, runs experiments to verify the implementation, and
reports results honestly. You operate autonomously through a tool-use loop.
The direction of each idea is fixed. Your engineering judgment determines
how best to implement it within the target codebase.

\smallskip
\noindent\textbf{System.}~All text output outside of tool use is shown to
the user. Tool results may contain data from external sources. If you
suspect a tool result contains a prompt injection attempt, flag it before
continuing. Prior messages are compressed automatically as the conversation
approaches context limits.

\smallskip
\noindent\textbf{Doing Tasks.}~Read files before modifying them. Do not
create new files unless absolutely necessary. Diagnose failure causes
before switching tactics. Do not add features, refactors, docstrings,
comments, or type annotations beyond what the assigned idea requires.
Verify that the implementation works before reporting completion.

\smallskip
\noindent\textbf{Executing Actions with Care.}~Evaluate the reversibility
of each action before proceeding. Local, reversible operations (file edits,
running tests) may be taken freely. Destructive or hard-to-reverse operations
(deleting branches, forced resets, overwriting uncommitted changes) require
careful consideration. Do not use destructive actions as shortcuts or
bypass safety checks.

\smallskip
\noindent\textbf{Experiment Workflow.}~You are working in an isolated git
worktree branched from the current trunk HEAD. All code changes happen on
this experiment branch. Do not commit to main or master. Baseline scores
are provided in the Evaluation Info section of your prompt. Run the
evaluation command on the unmodified codebase only if no baseline scores
are provided. Save experiment results to
\texttt{results/\{node\_id\}-\{description\}/} and commit only small
diagnostic files using \texttt{git add -f}. Use \texttt{RunTraining} for
any training or evaluation command that takes more than five minutes.
Do not use polling loops. At the end, provide a concise report with
the following sections: Idea, Changes, Implementation Choices,
Baseline vs.\ Result, Analysis, and Insights.

\smallskip
\noindent\textbf{Critical Rules.}~The idea direction is non-negotiable.
You may not substitute a fundamentally different approach. Implementation
choices (specific architecture, hyperparameters, code placement) are yours.
Report significant implementation choices explicitly. If the idea
underperforms after a good-faith implementation, that is the finding to report.
\end{promptbox}

\subsection{Algorithm Workflow}
\label{app:algorithm-flow}

Algorithm~\ref{alg:appendix-htr} expands the HTR pseudocode from the main
paper (Algorithm~\ref{alg:htr}) with implementation-level detail. Notation
follows the main paper: $\iota_n$ denotes the insight recorded at node $n$,
$b_n$ the git branch reference, $s_n$ the $\Eval_\dev$ score, $r_n$ the
factual result, and $\iota_{\mathrm{anc}(n)}$ the concatenated insights on
the path from $n_0$ to $n$'s parent.

\begin{algorithm}[!t]
\small
\DontPrintSemicolon
\SetKwInOut{Input}{Input}\SetKwInOut{Output}{Output}
\SetKwProg{Fn}{Procedure}{:}{}
\SetKwFunction{Exec}{Executor}
\caption{Hypothesis Tree Refinement (HTR) with expanded implementation detail.
  The coordinator owns $\Tree$, and each \textsc{Executor} owns one worktree.}
\label{alg:appendix-htr}
\Input{$\mathcal{P}=(\Mat_0,\Obj,\Eval_{\dev},\Eval_{\test})$, budget $B$, branching $k$, parallelism $P$}
\Output{best artifact $\Mat^\star$, annotated hypothesis tree $\Tree$, and run summary report}

\tcp{Initialization}
init $\Tree=(\{n_0\},\emptyset)$, $b_{n_0}\leftarrow\Mat_0$, $\Mat_{\mathrm{best}}\leftarrow\Mat_0$\;
run $\Eval_{\dev}(\Mat_0)$ and record baseline score and eval command in $\Tree.\mathrm{meta}$ via \texttt{TreeSetMeta}\;

\tcp{Main coordinator loop}
\While{$B$ left $\wedge$ pending leaves exist $\wedge$ no stop signal}{
  \tcp{Step 1: OBSERVE, build constraint view}
  $\mathcal{V}\leftarrow\textsc{Observe}(\Tree,\Mat_{\mathrm{best}})$
    \tcp*{shape, root insight, pruned lessons, validated findings}

  \tcp{Step 2: IDEATE, skill-gated hypothesis proposal}
  $\mathcal{V}_c \leftarrow \texttt{TreeView}(\mathrm{format}{=}\texttt{constraints})$
    \tcp*{hard constraints: no re-tread of pruned directions}
  load \texttt{idea\_drafting} skill via \texttt{LoadSkill}
    \tcp*{mandatory gate: must precede any candidate proposal}
  $p\leftarrow\textsc{Select}_{\textsf{parent}}(\mathcal{V})$\;
  \ForEach{surviving candidate after Fatal-Flaw Scan}{
    attach pending child $n^{(i)}$ with hypothesis $h^{(i)}$ = (Mechanism, Hypothesis, Observable, Conflicts)\;
    $\iota_{\mathrm{anc}(n^{(i)})}\leftarrow$ insights on $\textsf{path}(n_0\to p)$
      \tcp*{injected into Executor prompt}
    prune \texttt{idea\_drafting} scratch from coordinator context
      \tcp*{elide skill body + reasoning post-TreeAddNode}
  }

  \tcp{Step 3: SELECT frontier for parallel dispatch}
  $L\leftarrow$ up to $P$ pending leaves under $\textsc{Select}_{\textsf{frontier}}(\mathcal{V})$\;

  \tcp{Step 4: DISPATCH, parallel Executor dispatch}
  $\{(s_n,r_n,\iota_n,b_n)\}_{n\in L}\leftarrow\textbf{parallel }\Exec(h_n,\iota_{\mathrm{anc}(n)},\Mat_{\mathrm{best}})$\;

  \tcp{Step 5: UPDATE, write back and propagate}
  \ForEach{$n\in L$, $a\in\textsf{path}(n_0\to n)$}{
    write back $(s_n,r_n,\iota_n,b_n)$ to node $n$ in $\Tree$ and set $n.\mathrm{status}\leftarrow\texttt{done}$\;
    $\iota_a\leftarrow\textsc{Abstract}(\{\iota_c\}_{c\in\textsf{ch}(a)})$
      \tcp*{propagate insights upward to root}
  }
  check convergence: inject intervention if $\geq w$ consecutive non-improving experiments\;

  \tcp{Step 6: DECIDE, merge gate or prune}
  $n^\dagger\leftarrow\arg\max_{n\in L}\,s_n$\;
  \If{$s_{n^\dagger}$ exceeds best score by $\geq\theta$}{
    create detached worktree at $b_{n^\dagger}$ and run $\Eval_{\test}$ with template substitution\;
    \lIf{$S_{\test}(b_{n^\dagger})>S_{\test}(\Mat_{\mathrm{best}})$}{
      $\Mat_{\mathrm{best}}\leftarrow\textsf{merge}(b_{n^\dagger})$ and update $\Tree.\mathrm{meta}.\mathrm{trunk\_score}$
    }
  }
  prune subtrees falsified by $\{\iota_n\}_{n\in L}$ and persist $\Tree$ to JSON + Markdown\;
}
run $\Eval_{\test}(\Mat_{\mathrm{best}})$ and record \texttt{test\_trunk\_score} and \texttt{test\_baseline\_score}\;
\Return{$\Mat^\star\leftarrow\Mat_{\mathrm{best}}$, $\Tree$, run summary}\;

\BlankLine
\Fn{\Exec{$h_n,\,\iota_{\mathrm{anc}(n)},\,\Mat_{\mathrm{best}}$}}{
  branch name $\leftarrow \textsf{slug}(\mathrm{node\_id}) + \textsf{slug}(h_n) + \mathrm{SHA1}(h_n)_{[:8]}$\;
  create worktree $W_n$ in $\texttt{/tmp/}$ from current best branch HEAD on a fresh branch\;
  inject eval command (with \texttt{\{cwd\}$\to W_n$}, \texttt{\{node\_id\}$\to n$}) and $\iota_{\mathrm{anc}(n)}$ into prompt\;
  \Repeat{run ok $\wedge$ $h_n$-path exercised, or turn cap reached}{
    $\Delta\leftarrow\textsf{Implement}(h_n,\iota_{\mathrm{anc}(n)},W_n)$\;
    $(s_n,r_n)\leftarrow\Eval_{\dev}(\textsf{apply}(\Delta,W_n))$
      \tcp*{repair $\Delta$ only, direction $h_n$ is fixed}
  }
  filter $\Delta$: commit implementation files only, skip logs/checkpoints/caches\;
  remove worktree directory and retain branch $b_n$ for later merge gate\;
  \Return{$(s_n,\;r_n,\;\textsc{Distill}(h_n,\Delta,r_n),\;b_n)$}\;
}
\end{algorithm}

\subsection{Key Design of \arbor{} Framework}
\label{app:framework-key-design}
General-purpose coding agents such as Codex and Claude Code are built for
general-purpose software tasks: they chain tool calls on a single working tree
to edit, test, and fix code against a goal that is already well specified.
\arbor{} instead targets the auto-research setting, and we make a series of
engineering choices that specialize the agent for it so that the system can
flexibly adapt to different research needs and manage many experiments over a
long horizon. We group its key engineering designs into four areas: (i)~hypothesis-tree management,
(ii)~experiment management, (iii)~long-horizon operation, and (iv)~functional
extensibility through skills and plugins.

\paragraph{Hypothesis-tree management.}
Unlike a code agent whose memory is the linear chat transcript, \arbor{}
externalizes its research state into an explicit hypothesis tree (the
\emph{idea tree}), an in-memory object that is the single authoritative record
of a run. Each node holds a hierarchical dotted address (e.g.\ \texttt{ROOT},
\texttt{1}, \texttt{1.1}) that encodes its path from the root, its
\texttt{parent\_id} and \texttt{children\_ids}, a once-written
\texttt{hypothesis}, a \texttt{status} field tracing the lifecycle
\texttt{pending}\,$\to$\,\texttt{running}\,$\to$\,\texttt{done}\,$\to$\,%
\{\texttt{merged},\,\texttt{pruned}\}, a development \texttt{score}, a factual
\texttt{result}, a distilled \texttt{insight}, and a \texttt{code\_ref} branch
pointer to the artifact. Depth-1 nodes are broad directions and deeper nodes are
concrete refinements, so edges encode hypothesis refinement rather than
chronological actions. The coordinator never touches the raw structure; it
operates the tree only through a small set of typed tools
(\texttt{TreeAddNode}, \texttt{TreeUpdateNode}, \texttt{TreePrune},
\texttt{TreeSetMeta}, \texttt{TreePropagate}) and read projections
(\texttt{TreeView}). Storing an experiment is thus a controlled mutation: the
executor's structured report is parsed and written into the node's
\texttt{score}/\texttt{result}/\texttt{insight}/\texttt{code\_ref} fields, and
the tree is serialized to JSON (and rendered to Markdown) after \emph{every}
mutation. Crucially, when a node finishes \arbor{} \emph{backpropagates} its
insight: \texttt{propagate\_insights} walks from the node's parent to the root
and, at each ancestor, an LLM synthesizes the insights of that ancestor's
children into a concise ($<$200-word) summary. Leaf insights describe concrete
implementations, parent insights summarize families of interventions, and the
root insight maintains a global understanding of the problem. This layerwise
abstraction lets the coordinator reason at the right granularity without
rereading raw logs. The concrete schema and persistence format are detailed in
Section~\ref{app:idea-tree-storage}.

\paragraph{Experiment management.}
Each pending node is executed by an executor dispatched through
\texttt{RunSubagent} or \texttt{RunSubagentParallel}. Rather than editing the
shared working tree in place, \arbor{} creates a fresh git worktree branched
from the current trunk \texttt{HEAD} for each candidate, so every hypothesis gets
a clean, independently recoverable experimental boundary and several executors
can edit overlapping files concurrently without corrupting the trunk or one
another. At launch the executor is injected with the assigned hypothesis, the
ancestor insights along its path, and the task
objective, development evaluator $\Eval_\dev$, held-out evaluator $\Eval_\test$,
metric direction, data split, and baseline score stored in the tree metadata
via \texttt{TreeSetMeta}. Its permission boundary is deliberately narrow: the
executor's tools (Table~\ref{tab:sub-agent-tools}) act \emph{only} within its
own worktree. It cannot read the trunk, inspect sibling branches, mutate the
tree, or run $\Eval_\test$. It may repair implementation bugs and choose
reasonable engineering details, but it may not swap the assigned hypothesis for
a different research direction, which keeps the returned evidence attributable
to the node actually tested. The executor reports back four separated fields:
the development score $s_n$, a factual \texttt{result} $r_n$ (raw observations
such as errors, curves, and metric breakdowns), a distilled \texttt{insight}
$\iota_n$ (the causal lesson), and the branch reference $b_n$. The coordinator
auto-extracts these, updates the node, and decides admission: a result counts as
a \emph{useful gain} only when it improves $\Eval_\dev$ over the trunk by at
least the merge threshold, which triggers the \texttt{GitMergeBranch} held-out
gate that re-runs $\Eval_\test$ in a separate detached worktree and promotes the
artifact only if it strictly beats the current best. Every other outcome is
treated as \emph{failure evidence} rather than noise: the node's insight and, if
the direction is abandoned, its \texttt{TreePrune} reason are aggregated into
the constraint view (\texttt{TreeView(format="constraints")}) that conditions
the next \textsc{Ideate} step, so negative results actively narrow the search
space. Tool-level details of dispatch, score extraction, template-variable
substitution (\texttt{\{cwd\}}, \texttt{\{node\_id\}}), and artifact filtering
are given in Section~\ref{app:agent-tools-params}.

\paragraph{Long-horizon operation.}
A single auto-research run can span hundreds of turns and many hours of
evaluation, far beyond one context window, so \arbor{} adds explicit mechanisms
to keep the search productive instead of stalling on early failures or chasing
noisy evaluation swings. First, because all durable state lives in the persisted
idea tree rather than the transcript, the run survives crashes, agent restarts,
and context compression; post-commit context pruning further elides spent
\textsc{Ideate} scratch work and loaded skill bodies once a candidate is
committed, bounding context growth. Second, long training or evaluation commands
are routed through \texttt{RunTraining}, which blocks until completion or timeout
while continuously capturing partial metrics, progress logs, and checkpoints, so
even a run that times out at 80\% of its epochs returns actionable evidence
instead of a silent failure. Third, a \emph{convergence detector} monitors
recent score velocity over a sliding window and counts consecutive
non-improving experiments, escalating through \texttt{warn}, \texttt{paradigm\_shift},
and \texttt{stop} signals; it also flags \emph{parent exhaustion} when a parent's
recent children all fail to beat the trunk, prompting the coordinator to
summarize the failure pattern and open a fresh depth-1 direction rather than
over-exploring a dead branch. A meaningful-improvement threshold prevents a
single noisy uptick from resetting this signal, balancing premature stagnation
against unbounded exploration.

\paragraph{Functional extensibility (skills and plugins).}
To remain adaptable across research domains without changing the core loop,
\arbor{} exposes two extension surfaces. \emph{Skills} are markdown documents
with YAML frontmatter, discovered by a registry from both the built-in directory
and a project-local \texttt{.research\_agent/skills/} override, and loaded on
demand via \texttt{LoadSkill}. The \textsc{Ideate} protocol, for instance, loads
\texttt{idea\_drafting}, \texttt{first\_principles\_probe}, and
\texttt{fatal\_flaw\_scan} before proposing candidates, then prunes their bodies
from context after each commit, so reasoning guidance is injected just-in-time
and does not permanently inflate context. \emph{Plugins} are YAML domain
adapters that specialize the system declaratively rather than through code: each
plugin can inject domain guidance at six prompt points (coordinator/executor
init, ideate, decide, preamble, and workflow), declare an evaluation contract,
mark protected paths and required outputs for the merge guard, override runtime
configuration through named profiles, and tune convergence thresholds. For
example, the \texttt{mle\_kaggle} plugin configures a Kaggle/MLE-bench task with
its metric direction, evaluation command, protected data paths, and time-budget
profiles entirely in YAML. Together, skills and plugins let \arbor{} adapt to new
artifacts and evaluation regimes while keeping the hypothesis-tree machinery and
agent contracts unchanged.

\subsection{Agent Tools and Hyperparameter Settings}
\label{app:agent-tools-params}

\subsubsection{Coordinator Tools}
\label{app:meta-agent-tools}

The coordinator's tool set reflects a strict division of labor: it may read
any file in the repository and inspect the tree in any projection, but it
may never edit code directly. All implementation work is delegated to
executors via \texttt{RunSubagent} or \texttt{RunSubagentParallel}. This
design is intentional. If the coordinator could edit code directly, it
would be tempted to make small local fixes without creating a new tree
node, which would break the invariant that every code change is traceable
to a hypothesis. Enforcing the edit boundary at the tool level makes this
invariant a system property rather than a behavioral expectation.

Table~\ref{tab:meta-agent-tools} lists the full tool set. The tree tools
(\texttt{TreeView}, \texttt{TreeAddNode}, \texttt{TreeUpdateNode},
\texttt{TreePrune}, \texttt{TreeSetMeta}, \texttt{TreePropagate}) are the
primary interface through which the coordinator manages the shared research
state. The dispatch tools (\texttt{RunSubagent},
\texttt{RunSubagentParallel}) are the boundary through which the coordinator
hands off implementation to executors and receives structured evidence back.
The merge tool (\texttt{GitMergeBranch}) is the only path through which a
candidate branch can be promoted to the current best, enforcing the held-out gate at the
tool level rather than relying on the coordinator to remember to verify
before merging. Table~\ref{tab:meta-agent-hyperparams} reports the key
hyperparameters used across all experiments.

\begingroup
\footnotesize
\setlength{\tabcolsep}{3pt}
\renewcommand{\arraystretch}{1.02}
\setlength{\LTpre}{4pt}
\setlength{\LTpost}{6pt}
\begin{xltabular}{\linewidth}{>{\raggedright\arraybackslash}p{3.8cm}>{\raggedright\arraybackslash}X}
\caption{Tools available to the coordinator.}
\label{tab:meta-agent-tools}\\
\toprule
\textbf{Tool} & \textbf{Description} \\
\midrule
\endfirsthead
\caption[]{Tools available to the coordinator (continued).}\\
\toprule
\textbf{Tool} & \textbf{Description} \\
\midrule
\endhead
\midrule
\multicolumn{2}{r}{\footnotesize continued on next page}\\
\endfoot
\bottomrule
\endlastfoot
\texttt{TreeView} & Inspect the idea tree in five formats: \texttt{compact} (status overview), \texttt{full} (Markdown rendering), \texttt{node} (single-node detail), \texttt{pending} (pending leaf list), and \texttt{constraints} (aggregated pruned lessons and validated findings used as hard constraints in the IDEATE phase). \\
\texttt{TreeAddNode} & Add a child node to the tree with a four-field hypothesis block (Mechanism, Hypothesis, Observable, Conflicts). Triggers IDEATE context pruning after each commit. \\
\texttt{TreeUpdateNode} & Update mutable fields of an existing node (\texttt{status}, \texttt{insight}, \texttt{score}, \texttt{result}, \texttt{code\_ref}, \texttt{related\_work}). \\
\texttt{TreePrune} & Mark a node and its subtree as pruned with a written reason. The reason is aggregated into the constraints block for future IDEATE rounds. \\
\texttt{TreeSetMeta} & Write evaluation metadata to the tree root: evaluation commands (\texttt{eval\_cmd}, \texttt{eval\_cmd\_test}), dataset paths, baseline score, best score, and test scores. Automatically injected into every Executor's prompt. \\
\texttt{TreePropagate} & Re-propagate insights from a node upward to the root after manual corrections via \texttt{TreeUpdateNode}. \\
\texttt{RunSubagent} & Dispatch a single Executor to implement and evaluate a pending tree node. Creates an isolated git worktree from trunk HEAD, injects ancestor insights and evaluation metadata, auto-extracts results from the structured report, and updates the tree node. \\
\texttt{RunSubagentParallel} & Dispatch two to four Executors concurrently on independent tree nodes. Each Executor runs in its own isolated worktree. \\
\texttt{GitMergeBranch} & Validate and promote a candidate branch to the current best. Creates a temporary detached worktree at the source branch, runs $\Eval_\test$ with score extraction, and merges only if the validated test score exceeds the current best score. Protected branches (main, master) cannot be merge targets. \\
\texttt{Bash} & Execute shell commands for read-only codebase inspection and environment queries. The coordinator never uses Bash to edit code. \\
\texttt{FileRead} & Read the contents of a file within the target repository. \\
\texttt{Grep} & Search for a text pattern across repository files. \\
\texttt{Glob} & Find files matching a glob pattern within the repository. \\
\texttt{LoadSkill} & Load a skill document on demand from the registry. In the IDEATE protocol, \texttt{idea\_drafting}, \texttt{first\_principles\_probe}, and \texttt{fatal\_flaw\_scan} are loaded before proposing candidates. Loaded content is pruned from context after each \texttt{TreeAddNode} commit. \\
\texttt{SearchIdeaContext} & Dispatch a SearchAgent in the background to annotate a tree node with related work. Returns immediately. The SearchAgent runs concurrently with ongoing IDEATE and dispatch work and writes a structured annotation (summary, novelty assessment, related papers) to \texttt{node.related\_work} when finished. \\
\texttt{SearchIdeaContextParallel} & Dispatch multiple background SearchAgents concurrently for a list of node IDs. \\
\texttt{SearchStatus} & Return the count of in-flight background SearchAgent tasks. \\
\end{xltabular}
\endgroup

\begin{table}[!htbp]
\centering
\small
\setlength{\tabcolsep}{4pt}
\renewcommand{\arraystretch}{1.08}
\caption{Key hyperparameters for the \arbor{} coordinator and executors.}
\label{tab:meta-agent-hyperparams}
\begin{tabularx}{\linewidth}{>{\raggedright\arraybackslash}p{4.8cm}>{\raggedright\arraybackslash}p{2.4cm}>{\raggedright\arraybackslash}X}
\toprule
\textbf{Parameter} & \textbf{Default} & \textbf{Description} \\
\midrule
Backbone model & Claude Opus 4.6 & LLM for both coordinator and executors (configurable per role) \\
Max cycles & 20 & Hard cap on total completed experiments \\
Max coordinator turns & 500 & Maximum ReAct turns for the persistent coordinator loop \\
Executor max turns & 50 & Maximum turns per executor \\
Executor timeout & 48\,h & Wall-clock limit per executor \\
Context window & 200\,K tokens & Coordinator context limit \\
Compression threshold & 0.80 & Compress context when it reaches 80\% of the context window \\
Kept recent turns & 20 & Number of most recent turns preserved after each compression \\
Merge threshold & 5.0\% & Minimum $\Eval_\dev$ improvement required to invoke the merge gate \\
$\Eval_\test$ evaluation retries & 1 & Additional held-out evaluation attempts after a transient failure \\
Convergence window & 5 & Sliding window size for score velocity computation \\
Convergence warn / stop & 3 / 8 & Consecutive non-improving experiments to trigger warning / stop \\
\bottomrule
\end{tabularx}
\end{table}

\subsubsection{Executor Tools}
\label{app:sub-agent-tools}

The executor's tool set is designed around a single constraint: every action
must be local to its worktree. The executor has no tools for reading the
current best, inspecting sibling branches, or modifying the shared tree. This
isolation is not merely a safety measure. It is what allows multiple
executors to run concurrently on overlapping codebases without coordination
overhead. Each executor operates as if it has an exclusive copy of the
repository, because its worktree is literally a separate checked-out
directory that shares the git object store but maintains its own HEAD and
index. The \texttt{RunTraining} tool deserves special mention: for experiments
that involve multi-hour training runs, using \texttt{Bash} with a timeout
would either cut the run short or block the executor for an unbounded
duration. \texttt{RunTraining} instead blocks until the command terminates
or the configured timeout is reached, continuously capturing intermediate
metrics, progress logs, and checkpoints so that even a timed-out run
returns actionable partial evidence. This matters in practice: a training
run that completes 80\% of its epochs before timing out often reveals enough
about a direction's trajectory to inform a principled prune or continuation
decision, which is more useful than a silent timeout that leaves the
coordinator without evidence.

\begin{table}[!htbp]
\centering
\small
\setlength{\tabcolsep}{4pt}
\renewcommand{\arraystretch}{1.08}
\caption{Tools available to each executor.}
\label{tab:sub-agent-tools}
\begin{tabularx}{\linewidth}{>{\raggedright\arraybackslash}p{2.8cm}>{\raggedright\arraybackslash}X}
\toprule
\textbf{Tool} & \textbf{Description} \\
\midrule
\texttt{Bash} & Execute shell commands in the worktree with configurable per-call timeouts (default 600\,s, maximum 86{,}400\,s). \\
\texttt{RunTraining} & Execute a long-running training or evaluation command and block until completion or timeout. Automatically captures partial metrics, progress logs, and checkpoints. Supports staged budgets (smoke, pilot, full) with configurable wall-times. Maximum timeout is 604{,}800\,s (seven days). \\
\texttt{FileRead} & Read the full contents of a file within the worktree. \\
\texttt{FileEdit} & Edit a file by replacing an exact string match with new content, enabling surgical edits that minimize unintended side effects. \\
\texttt{FileWrite} & Write the complete contents of a file, creating or overwriting it. \\
\texttt{Grep} & Search for a text or regex pattern across files in the worktree. \\
\texttt{Glob} & Find files matching a glob pattern within the worktree. \\
\texttt{SubAgent} & Spawn a nested executor with a custom prompt for modular decomposition of complex implementation tasks. The nested agent inherits the same tool set and worktree context. \\
\bottomrule
\end{tabularx}
\end{table}

\subsubsection{Evaluation and Merge Tools}
\label{app:evaluation-merge-tools}

A recurring challenge in multi-agent systems is ensuring that evaluation
commands measure the right code. In \arbor{}, each executor runs in a
different directory, on a different branch, with potentially different data
locations. A naively hardcoded evaluation command would silently evaluate
the wrong artifact. The template variable system addresses this by requiring
the coordinator to specify evaluation commands using two placeholders:
\texttt{\{cwd\}} is substituted with the executor's worktree path before
the command reaches the executor's prompt, and \texttt{\{node\_id\}} is
substituted with the tree node identifier so that results from concurrent
experiments are written to distinct output locations and never overwrite
one another. The coordinator is also explicitly prohibited from hardcoding
absolute paths in evaluation commands. Only \texttt{\{cwd\}}-relative
paths are permitted.

Score extraction from evaluator output follows a two-stage pipeline
that handles the diversity of real evaluation scripts. Many harnesses
already emit a structured JSON block with a \texttt{score} key. The system
attempts to parse this first. If no valid JSON block is found, as is
common with training scripts that report loss or accuracy in free-form
log lines, a secondary LLM call reads the full command output and
extracts the primary metric as a percentage. The distinction between a
primary metric and ancillary logging is task-specific, so the LLM call
uses a concise system prompt that instructs it to return only a single
JSON object and to pick the most prominent performance figure if multiple
metrics appear.

The held-out evaluator $\Eval_\test$ is architecturally separated from
the development loop. Executors are instructed never to run $\Eval_\test$.
It is accessible only through the \texttt{GitMergeBranch} tool, which
creates a fresh detached worktree at the candidate branch's HEAD, applies
the same template substitution, and runs \texttt{eval\_cmd\_test} in
complete isolation from the main repository. Configurable retry logic
with exponential backoff handles transient infrastructure failures such as
gpu allocation timeouts or flaky network reads. The merge is admitted only
if the extracted test score strictly exceeds the current best score under
the metric direction specified in the task. This two-worktree design uses one
worktree for development evaluation inside the executor and one for held-out
evaluation inside the merge gate. It ensures that neither the executor nor
the coordinator
ever evaluates the candidate artifact in the same directory as the current best,
eliminating any risk of result leakage between branches.

Before a worktree branch is committed, \arbor{} applies artifact filtering
to separate implementation changes from generated outputs. The system
inspects the full worktree diff and classifies changed paths as either
implementation files or artifacts. Raw logs, model checkpoints,
cache directories, generated data, and large binary files are excluded
from the commit, keeping experiment branches compact and ensuring that
subsequent three-way merges into the current best branch involve only meaningful code
differences. This filtering is conservative by design: a borderline file
is included rather than excluded, because a missing implementation file
breaks the merge, whereas a spurious large file only adds storage overhead.

\subsection{Idea Tree Data Structure and Storage}
\label{app:idea-tree-storage}

The implementation stores \arbor{}'s hypothesis tree as an idea-tree object.
This object is the primary durable data structure of a run: it records the
research frontier, completed experiments, rejected directions, accepted
artifacts, and reusable insights in one inspectable state.

\paragraph{Logical schema.}
The root node anchors the initial artifact and task contract. Depth-1 nodes
represent broad research directions, while deeper nodes represent concrete
refinements, alternatives, or corrections under their parent. Edges therefore
encode hypothesis refinement rather than chronological agent actions. During a
run, pending leaves form the executable frontier; completed, merged, and pruned
nodes remain in the tree as evidence for future proposal and selection.
Each node carries the fields listed in Table~\ref{tab:node-fields}.

\begin{table}[ht]
\centering
\small
\setlength{\tabcolsep}{4pt}
\renewcommand{\arraystretch}{1.10}
\caption{Fields of a single node in the idea tree.
The first four fields form the structural skeleton;
the remaining fields are populated progressively as the node
moves through its status lifecycle.}
\label{tab:node-fields}
\begin{tabularx}{\linewidth}{>{\raggedright\arraybackslash}p{2.4cm}%
>{\raggedright\arraybackslash}p{2.8cm}>{\raggedright\arraybackslash}X}
\toprule
Field & Type & Description \\
\midrule
\texttt{id} & string
  & Hierarchical address encoding position in the tree
    (e.g.\ \texttt{"ROOT"}, \texttt{"1"}, \texttt{"1.1"}).
    The dot-separated prefix gives the unique path from root. \\
\texttt{parent\_id} & string\,\textbar\,null
  & Identifier of the parent node. \texttt{null} for the root. \\
\texttt{children\_ids} & list[string]
  & Ordered identifiers of child nodes, appended as the
    coordinator expands the tree during \textsc{Ideate}. \\
\texttt{depth} & int
  & Integer depth: $0$~=~root, $1$~=~direction node,
    $2$~=~branch node.
    The maximum depth is governed by \texttt{max\_depth}. \\
\addlinespace
\texttt{hypothesis} & string
  & Research hypothesis assigned before execution.
    Written once by the coordinator during \textsc{Ideate} and
    never modified thereafter. \\
\texttt{status} & \{pending, running,\newline done, merged, pruned\}
  & Status lifecycle: \texttt{pending}~$\to$~\texttt{running}~$\to$~%
    \texttt{done}, then \texttt{done}~$\to$~\texttt{merged}
    if the merge gate passes, or \texttt{done}~$\to$~\texttt{pruned}
    otherwise. \\
\texttt{score} & float\,\textbar\,null
  & Absolute scalar score on $\Eval_{\dev}$ as reported by the executor.
    Null until the node has been executed and scored. \\
\texttt{result} & string
  & Factual record of the experiment outcome, written by the executor
    as a direct observational report before any interpretation is applied. \\
\texttt{insight} & string
  & Structured lesson extracted by \textsc{Distill} from the outcome,
    then backpropagated and aggregated at each ancestor up to the root. \\
\texttt{code\_ref} & string\,\textbar\,null
  & Git branch name pointing to the implementation artifact produced
    by the executor.
    Null if no artifact was committed. \\
\bottomrule
\end{tabularx}
\end{table}

\paragraph{Persistent storage.}
The idea tree is serialized to JSON after every controlled mutation and serves
as the sole authoritative record of research state accumulated during a run.
No hypothesis, score, artifact reference, or lesson exists outside this
structure. The same state is also rendered to Markdown for dashboards and human
inspection. The JSON representation is consumed by tools, while the Markdown
view provides a readable account of the research process. Since every mutation
is persisted immediately, the run can recover from crashes, agent restarts, and
context-window compression without relying on conversational memory.

\paragraph{Agent-facing access.}
Agents do not edit the raw JSON file directly. The coordinator accesses the
tree through controlled read views and mutation tools: compact frontier views,
single-node views, pending-leaf views, and the constraint view used before
ideation; plus mutations for adding nodes, updating lifecycle fields, pruning
subtrees, setting metadata, and propagating insights. Executors receive only the
assigned hypothesis, the research contract, and ancestor insights. This access
pattern keeps the tree as a shared state object rather than a low-level log of
tool traces.

\FloatBarrier

\section{Details of the AO Test Suite}
\label{app:ao-suite-details}
\label{app:benchmark-details}

\subsection{Optimizer Design}
\label{app:benchmark-optimizer-design}

NanoGPT-Bench~\citep{nanogptbench} Track~3 is a collaborative benchmark
for discovering efficient neural network optimizers.
Unlike the main NanoGPT speedrun, which minimizes wall-clock time by any
means, Track~3 targets \emph{step count}: methods that are slow in
wall-clock terms but reach the target loss in fewer steps are valid and
desirable.
All optimizer designs are evaluated on the same fixed architecture, dataset,
and training script, so the only lever available is the optimization
algorithm and its hyperparameters.

\paragraph{Baseline.}
We initialize from the official tuned Muon optimizer setting provided by the
benchmark authors.
This configuration applies Muon to transformer block weights and AdamW to
embeddings, the output projection, and scalar parameters (biases and gains).
With these tuned hyperparameters, the baseline reaches a FineWeb validation
loss of $\leq 3.28$ in 3{,}325 steps.
Selecting a well-tuned official baseline as the starting material is
intentional: it places the agent in a regime where naive hyperparameter
sweeps are unlikely to yield large gains, thereby testing the agent's ability
to identify and implement genuinely novel optimizer improvements rather than
harvesting easy wins from an undertuned starting point.

\paragraph{Development and test split.}
Agents iterate with the development evaluator throughout the search loop.
After the search completes, the selected optimizer is re-evaluated with two
held-out random seeds, and the reported test score is the average step count
across those runs.

\paragraph{Evaluator.}
The evaluator runs the benchmark's official evaluation script
\texttt{run\_eval.py}.
The script launches the training job, monitors validation loss, and
terminates as soon as \texttt{val\_loss} $\leq 3.28$ is first reached.
The score is the step count at termination, with lower being better.
If the target is never reached, a penalty score exceeding 7{,}000 is
assigned.
All evaluation runs are executed on four NVIDIA A100-80\,GB GPUs using the
official benchmark evaluation code.

\paragraph{Agent instruction.}
The agent is initialized with the following task description.

\begin{promptbox}[title=\textbf{Optimizer Design Task Instruction}]
Improve the training-step efficiency of the NanoGPT optimizer on FineWeb.
The training script \texttt{train\_gpt\_simple.py} trains a 124\,M-parameter
GPT-2 on FineWeb using Muon for transformer weights and AdamW for embeddings,
output projection, and scalar parameters.

\smallskip
\noindent\textbf{Goal.}~Minimize the number of training steps required to
reach \texttt{val\_loss}~$\leq 3.28$.
The current baseline achieves this target in 3{,}325 steps.

\smallskip
\noindent\textbf{Evaluation.}~Run \texttt{python run\_eval.py} from the
project directory.
The script launches training, monitors validation loss, and terminates at the
first step where \texttt{val\_loss}~$\leq 3.28$.
The score is the step count at termination, with lower being better.
If the target is never reached, a penalty score exceeding 7{,}000 is assigned.

\smallskip
\noindent\textbf{Constraints.}~Only \texttt{train\_gpt\_simple.py} may be
modified.
The dataset, batch size, and model architecture must remain unchanged.
Adjusting \texttt{train\_steps} in isolation is not a valid improvement.
Multiple forward passes per step are not permitted.
\end{promptbox}

\subsection{Architecture Design}
\label{app:benchmark-architecture-design}

Architecture Design uses the \texttt{autoresearch} benchmark~\citep{autoresearch},
a compact LLM pretraining task designed for closed-loop research on a real
training codebase.
The agent receives a single-file training implementation and must improve the
final validation loss under a fixed wall-clock training budget.
Unlike Optimizer Design, where the model architecture is fixed and only the
optimizer is modified, this task exposes the full training recipe: model shape,
attention pattern, initialization, optimizer hyperparameters, learning-rate
schedules, batch sizing, and training-loop details may all be changed as long
as the benchmark evaluator and data preparation remain fixed.

\paragraph{Baseline.}
The initial material is the default \texttt{autoresearch} repository.
The main artifact is \texttt{train.py}, a single-GPU decoder-only Transformer
pretraining script derived from \texttt{nanochat}.
Data preparation, tokenization, validation-token selection, and evaluation
helpers live outside the editable artifact, primarily in \texttt{prepare.py}.
The development evaluator uses the fixed validation shard provided by the benchmark.
The default model is a roughly 50M-parameter Transformer, and the baseline
reports the final validation bits-per-byte after the fixed training run.

\paragraph{Development and test split.}
During search, agents optimize only against the development evaluator, which
runs the training script on the benchmark's fixed prepared data and reports the
final validation loss.
The held-out score reported in the main experiments is obtained by rerunning
the selected \texttt{train.py} with two held-out random seeds and averaging the
resulting final losses.
This separates the fast, single-run feedback used for hypothesis exploration
from the seed-averaged score used for final comparison.

\paragraph{Evaluator.}
The evaluator executes
\texttt{uv run train.py} from the benchmark repository.
At the end of training the script prints a structured summary including
\texttt{val\_bpb}, \texttt{peak\_vram\_mb}, \texttt{training\_seconds},
\texttt{num\_steps}, and \texttt{mfu\_percent}.
The primary score is the final \texttt{val\_bpb}, with lower values better.
The training script enforces a 300-second training budget; runs that crash,
time out, fail to fit in memory, or do not emit a parseable final
\texttt{val\_bpb} are treated as failed experiments.
The agent may use secondary diagnostics such as step count and peak memory to
interpret failures, but merge decisions are based on the loss metric.

\paragraph{Agent instruction.}
The agent is initialized with the following task description.

\begin{promptbox}[title=\textbf{Architecture Design Task Instruction}]
Follow \texttt{program.md} as the authoritative task specification for the
\texttt{autoresearch} repository.

\smallskip
\noindent\textbf{Goal.}~Minimize the validation bits-per-byte
(\texttt{val\_bpb}) reported by \texttt{uv run train.py}.
Lower is better.

\smallskip
\noindent\textbf{Evaluation.}~Run \texttt{uv run train.py} from the project
directory and extract the final \texttt{val\_bpb}, together with
\texttt{peak\_vram\_mb}, \texttt{training\_seconds}, \texttt{num\_steps},
and \texttt{mfu\_percent} for diagnostics.
The script has a fixed 5-minute training budget.
If a run crashes, runs out of memory, exceeds the allowed wall-clock budget, or
does not produce a parseable final loss, treat it as a failed experiment unless
there is an obvious implementation bug that can be corrected quickly.

\smallskip
\noindent\textbf{Constraints.}~Only \texttt{train.py} may be modified.
Architecture, hyperparameters, optimizer behavior, batch size, model size,
attention pattern, initialization, schedules, and training-loop details are in
scope.
Do not modify \texttt{prepare.py}, especially \texttt{TIME\_BUDGET},
\texttt{MAX\_SEQ\_LEN}, \texttt{EVAL\_TOKENS}, \texttt{make\_dataloader}, or
\texttt{evaluate\_bpb}.
Do not modify data files, tokenizer artifacts, \texttt{pyproject.toml},
\texttt{uv.lock}, or install new dependencies.
\end{promptbox}

\subsection{Terminal-Bench 2.0}
\label{app:benchmark-terminal-bench}

Terminal-Bench~2.0~\citep{terminalbench} is a benchmark for evaluating
terminal agents on realistic command-line tasks executed inside isolated
Docker containers.
Tasks span 16 categories, including software engineering, security,
scientific computing, data science, games, debugging, and others, across
easy, medium, and hard difficulty levels.
The full task pool contains 89 tasks.

\paragraph{Baseline.}
Prior harness-engineering studies may begin from codebase variants that
differ in interface extensibility, tool set, or prompt structure, creating
comparison confounds.
To ensure a fair and reproducible starting point we use the official
terminal-agent codebase (\texttt{terminus-2}) distributed with the
benchmark verbatim, without any modifications and without introducing
additional interfaces.
The backbone model is GPT-5.5.
The agent operates as a ReAct-style loop that issues tmux keystrokes to
a persistent terminal session.

\paragraph{Development and test split.}
We stratify the 89 tasks by difficulty and sample a 36-task development
set and a 53-task held-out test set.
The stratification ensures that both splits have balanced difficulty
coverage, preventing a scenario where the agent is evaluated on an
atypically easy or hard subset.
Agents iterate exclusively on the development set during the research
loop.
The test set is evaluated only once after the search completes, to
report held-out performance.

\paragraph{Evaluator.}
The evaluator is the official Harbor evaluation harness distributed with
Terminal-Bench~2.0.
Both development and test evaluations use 8 concurrent workers.
The evaluator reports pass rate as the fraction of tasks solved correctly,
with higher being better.
The baseline achieves 58.33\% on the development set (21 out of 36 tasks).

\paragraph{Agent instruction.}
The agent is initialized with the following task description.

\begin{promptbox}[title=\textbf{Terminal-Bench 2.0 Task Instruction}]
Improve the pass rate of a terminal agent on Terminal-Bench~2.0.
The current baseline achieves 58.33\% on the 36-task development set
using GPT-5.5.

\smallskip
\noindent\textbf{Goal.}~Maximize pass rate on the development split.
Iterate on the development split exclusively.
Evaluate the held-out test split only after achieving a meaningful
improvement on development.

\smallskip
\noindent\textbf{Evaluation.}~Development: \texttt{HARBOR\_N\_CONCURRENT=8 python3 run\_eval.py --data data/dev.json}.
Test: \texttt{HARBOR\_N\_CONCURRENT=8 python3 run\_eval.py --data data/test.json}.
The score is the fraction of tasks solved correctly, with higher being better.

\smallskip
\noindent\textbf{Constraints.}~Modifiable components include the system
prompt, per-task extra instructions, the agent subclass, the base agent
and its ReAct loop, response parsers, terminal session management, and
new files under the agent or prompts directories.
The evaluation harness, task data files, API configuration, and baseline
reference record must not be modified.
\end{promptbox}

\subsection{BrowseComp}
\label{app:benchmark-browsecomp}

BrowseComp~\citep{browsecomp} is a search-agent benchmark for
multi-step question answering.
The material to be optimized is not the benchmark data itself, but a minimal
ReAct-style browsing harness~\citep{react} that answers BrowseComp questions
using search and page-reading tools.
The task measures whether an autonomous research agent can improve the control
logic around a search agent while leaving the evaluator and question sets
fixed.

\paragraph{Baseline.}
The initial material is a simple ReAct-based search harness centered
on \texttt{single\_agent\_gpt.py}.
For each question, a \texttt{GPTAgent} runs a single ReAct trajectory with a
system prompt, issues web-search and page-visit tool calls, and returns a
short final answer.
The online BrowseComp setting uses \texttt{SearchTool} and \texttt{VisitTool};
the local-corpus tool variants are available in the codebase but are not the
primary path for this benchmark.
The baseline uses the simple-evals BrowseComp query template and grader
template, with the same model family used for the answering agent and the
grader in the reported runs.

\paragraph{Development and test split.}
The development split contains 50 BrowseComp questions and is used for all
iterative optimization.
The held-out test split contains 300 non-overlapping BrowseComp questions and
is reserved for final verification.
In the released harness, these splits are exposed as
\texttt{data/bc\_val.jsonl} and \texttt{data/bc\_test.jsonl}.
Agents may inspect and evaluate on the development split during the search
loop, but they may not edit either data file or use the test split for
iteration.

\paragraph{Evaluator.}
The development evaluator is
\texttt{uv run python run\_eval.py --data data/bc\_val.jsonl --workers 8 --run-name \{node\_id\}}.
The held-out evaluator uses the same entry point with
\texttt{data/bc\_test.jsonl} and a distinct run name.
The evaluator sends each question to the harness, collects the final answer,
normalizes it through the fixed BrowseComp answer-grading path, and reports
accuracy as \texttt{correct}/\texttt{total}.
Items with repeated execution errors or unparsable final answers are counted as
errors and do not contribute to the correct count.
The primary metric is accuracy, with higher values better.

\paragraph{Search and tool configuration.}
The harness exposes two browsing tools to the answering agent.
\texttt{SearchTool} accepts one or more search queries and returns compact web
search results, while \texttt{VisitTool} fetches and cleans page content for a
selected URL.
The tool names and argument schemas are fixed so that alternative harness
designs remain compatible with the same ReAct agent interface.
Agents may change the system prompt, rollout orchestration, response parsing,
candidate aggregation, and other harness code, but the evaluation script,
question files, grader template, and reference answers are fixed.

\paragraph{Transfer-test protocol.}
After optimization on BrowseComp, the resulting search harness is frozen and
evaluated directly on unseen search-agent tasks, including HLE and
DeepSearchQA, without additional task-specific optimization.
The same code path, search/visit tool interface, and answer-production protocol
are reused.
This transfer test checks whether the discovered harness changes improve
general browsing behavior rather than merely fitting the BrowseComp development
questions.

\paragraph{Agent instruction.}
The agent is initialized with the following task description.

\begin{promptbox}[title=\textbf{BrowseComp Task Instruction}]
Optimize BrowseComp search-agent accuracy starting from the simple baseline on
the current main branch.
Do not use prior optimized branches, cached scores, or old result directories
as the starting point.

\smallskip
\noindent\textbf{Goal.}~Maximize answer accuracy on the BrowseComp
development split.
Use development feedback to propose and evaluate harness changes, and reserve
the held-out test split for final verification.

\smallskip
\noindent\textbf{Evaluation.}~Development:
\texttt{uv run python run\_eval.py --data data/bc\_val.jsonl --workers 8 --run-name \{node\_id\}}.
Test:
\texttt{uv run python run\_eval.py --data data/bc\_test.jsonl --workers 8 --run-name \{node\_id\}\_test}.
The evaluator uses the fixed simple-evals BrowseComp prompt and LLM grader.
The score is accuracy, with higher values better.

\smallskip
\noindent\textbf{Constraints.}~The agent may modify the search harness,
including \texttt{single\_agent\_gpt.py}, prompts, agent-control logic,
response parsers, rollout strategies, and aggregation code.
Do not modify \texttt{run\_eval.py}, the development or test data files, API
configuration, the grader template, or cached reference records.
\end{promptbox}

\subsection{Search-Agent Data Synthesis}
\label{app:benchmark-search-agent-data}

Search-Agent Data Synthesis is a pipeline-optimization benchmark that
evaluates whether a research agent can improve a synthetic question-answering
generation system.
The system begins from Wikipedia-derived topic seeds and produces
BrowseComp-style information-retrieval questions.
A high-quality generated item must satisfy three properties simultaneously:
structural well-formedness, answerability by a search-enabled model, and
sufficient difficulty that the solver does not answer it reliably in a single
attempt.
The task therefore measures not only whether generated questions are
answerable, but also whether they require multi-step evidence gathering rather
than surface-level lookup.

\paragraph{Baseline.}
The initial material is the unmodified default data-synthesis pipeline
distributed with the benchmark.
The baseline pipeline consists of five stages: extracting candidate facts from
a topic seed, rating facts by information value, constructing a question whose
answer is one selected fact, obfuscating the topic with supporting constraints,
and applying structural verification.
The generator, solver, and judge use GPT-5.5 through an OpenAI-compatible
interface.

\paragraph{Development and test split.}
The benchmark provides frozen seed snapshots for both evaluation splits.
The development split contains 50 seeds and is used exclusively for iterative
optimization.
The held-out test split contains 100 seeds and is reserved for final
verification after the search loop completes.
The pipeline generates at most one QA item per seed.
Fixing the seed snapshots makes the benchmark self-contained and ensures that
measured improvements reflect changes to the synthesis mechanism rather than
changes to the data distribution.

\paragraph{Evaluator.}
The evaluator is the benchmark's fixed \texttt{run\_eval.py} entry point.
It loads the requested seed split, runs the synthesis pipeline, solves each
generated item four times with a web-search-enabled ReAct model, and reports
a JSON summary on the final line of standard output.
The primary metric is:
\[
\mathrm{score}
\;=\;
\frac{1}{N}\sum_{i=1}^{N}
\bigl(\mathrm{pass@4} - \mathrm{pass@1}\bigr),
\]
where $\mathrm{pass@1}$ is $1$ if the first attempt is judged correct, and
$\mathrm{pass@4}$ is $1$ if at least one of four attempts is correct.
Higher scores indicate questions that remain solvable in principle but are
not answered reliably on the first attempt.

\paragraph{Agent instruction.}
The agent is initialized with the following task description.

\begin{promptbox}[title=\textbf{Search-Agent Data Synthesis Task Instruction}]
Improve the quality of generated search-intensive QA items by optimizing
the data-synthesis pipeline.
The goal metric rewards items that are solvable under repeated attempts but
not answered reliably on the first attempt:
\[
\mathrm{score}
\;=\;
\mathrm{mean}(\mathrm{pass@4} - \mathrm{pass@1}).
\]

\smallskip
\noindent\textbf{Goal.}~Maximize the score on the development split.
Prefer mechanism-level improvements that preserve well-formedness and
solvability while increasing multi-step evidence gathering.

\smallskip
\noindent\textbf{Evaluation.}~Development: \texttt{python run\_eval.py --split dev}.
Test: \texttt{python run\_eval.py --split test}.
Iterate on the development split. Run the test split only after a
meaningful improvement.

\smallskip
\noindent\textbf{Constraints.}~Only the synthesis pipeline under
\texttt{pipeline/} may be modified, including pipeline steps, the async
runner, the step registry, pipeline configuration, and prompt files.
The scoring module, evaluation harness, benchmark configuration, and
frozen seed snapshots must not be modified.
\end{promptbox}

\subsection{Math-Reasoning Data Synthesis}
\label{app:benchmark-math-reasoning-data}
\label{app:math-case-study}

Math-Reasoning Data Synthesis is a pipeline-optimization benchmark that
evaluates whether a research agent can improve a system for generating
mathematical contest problems.
The pipeline starts from fixed AIME/AIMO/NuminaMath-style seeds and produces
original problems with exact integer answers in the range 0--999.
The research target is the data-construction mechanism itself: generated
problems must be valid, novel, diverse, mathematically well specified, and
calibrated so that a fixed reference solver does not answer them trivially.

\paragraph{Baseline.}
The initial material is the unmodified default synthesis pipeline distributed
with the benchmark.
The pipeline takes each seed, generates several candidate contest problems,
filters candidates for answer format, rationale consistency, held-out overlap,
and near-duplicate overlap, and evaluates surviving candidates with a fixed
solver.
The default generator uses GPT-5.5-mini and the reference solver uses GPT-5.5.
The answer policy is \texttt{integer\_0\_999}.

\paragraph{Development and test split.}
The benchmark uses locked seed files for both evaluation splits.
The development split contains 10 seeds with 5 generated candidates per seed,
yielding up to 50 candidates total.
The held-out test split contains 12 seeds with 8 generated candidates per seed,
yielding up to 96 candidates total.
Each seed specifies a mathematical topic, technique, and target difficulty.
Agents iterate exclusively on the development split during the research loop.
the test split is reserved for milestone and final verification.

\paragraph{Evaluator.}
The evaluator is the fixed \texttt{run\_eval.py} entry point.
It loads the requested split, runs the synthesis pipeline, applies validity
and novelty filters, evaluates each surviving candidate with the fixed solver,
and prints a JSON metrics object on the final line of standard output.
The primary metric is:
\[
\mathrm{score}
\;=\;
\frac{1}{N}
\sum_{i=1}^{N}
\bigl(\mathrm{pass@4} - \mathrm{pass@1}\bigr),
\]
where $N$ is the number of generated candidates before filtering,
$\mathrm{pass@1}$ indicates whether the first solver sample is correct,
and $\mathrm{pass@4}$ indicates whether at least one of four solver
samples is correct.
Filtered candidates are treated as zero.
This metric rewards problems that are solvable under repeated attempts but
not immediately solved on the first sample.

\paragraph{Agent instruction.}
The agent is initialized with the following task description.

\begin{promptbox}[title=\textbf{Math-Reasoning Data Synthesis Task Instruction}]
Improve a pipeline that generates AIME/NuminaMath-style contest problems
with exact integer answers in the range 0--999.
Generated problems must be valid, novel, diverse, and calibrated so that the
fixed reference solver fails on the first sample but succeeds within four
samples.
The primary metric is $\mathrm{mean}(\mathrm{pass@4} - \mathrm{pass@1})$ over all
generated candidates.

\smallskip
\noindent\textbf{Goal.}~Maximize the primary metric reported on the final
standard-output line of the evaluator.
Prefer mechanism-level improvements such as structured or parametric
generation, programmatic answer computation, and difficulty calibration.

\smallskip
\noindent\textbf{Evaluation.}~Development: \texttt{uv run python run\_eval.py --split dev}.
Test: \texttt{uv run python run\_eval.py --split test}.
Iterate on the development split. Run the test split only for milestone or
final verification.

\smallskip
\noindent\textbf{Constraints.}~Modifiable files include
\texttt{configs/pipeline.yaml}, \texttt{prompts/generate\_problem.md},
\texttt{src/math\_synth\_bench/baseline.py}, and new modules under
\texttt{src/math\_synth\_bench/}.
The benchmark harness, seed files, evaluation references, metrics module,
and verification module must not be modified.
\end{promptbox}

\FloatBarrier


\begin{thebibliography}{59}
\providecommand{\natexlab}[1]{#1}
\providecommand{\url}[1]{\texttt{#1}}
\expandafter\ifx\csname urlstyle\endcsname\relax
  \providecommand{\doi}[1]{doi: #1}\else
  \providecommand{\doi}{doi: \begingroup \urlstyle{rm}\Url}\fi

\bibitem[{Anthropic}(2025)]{claudecode}
{Anthropic}.
\newblock {Claude Code}.
\newblock \url{https://github.com/anthropics/claude-code}, 2025.
\newblock Agentic coding tool for terminal, IDE, and GitHub workflows. Accessed: 2026-06-02.

\bibitem[Chai et~al.(2025)Chai, Tang, Ye, Du, Zhu, Zhou, Wang, E, Zhang, Zhang, and Chen]{scimaster}
Jingyi Chai, Shuo Tang, Rui Ye, Yuwen Du, Xinyu Zhu, Mengcheng Zhou, Yanfeng Wang, Weinan E, Yuzhi Zhang, Linfeng Zhang, and Siheng Chen.
\newblock Scimaster: Towards general-purpose scientific {AI} agents, part i. x-master as foundation: Can we lead on humanity's last exam?
\newblock \emph{CoRR}, abs/2507.05241, 2025.
\newblock \doi{10.48550/ARXIV.2507.05241}.
\newblock \url{https://doi.org/10.48550/arXiv.2507.05241}.

\bibitem[Chan et~al.(2024)Chan, Chowdhury, Jaffe, Aung, Sherburn, Mays, Starace, Liu, Maksin, Patwardhan, Weng, and Madry]{mlebench}
Jun~Shern Chan, Neil Chowdhury, Oliver Jaffe, James Aung, Dane Sherburn, Evan Mays, Giulio Starace, Kevin Liu, Leon Maksin, Tejal Patwardhan, Lilian Weng, and Aleksander Madry.
\newblock Mle-bench: Evaluating machine learning agents on machine learning engineering.
\newblock \emph{CoRR}, abs/2410.07095, 2024.
\newblock \doi{10.48550/ARXIV.2410.07095}.
\newblock \url{https://doi.org/10.48550/arXiv.2410.07095}.

\bibitem[Chen et~al.(2026{\natexlab{a}})Chen, Chen, Chen, Zhao, Meng, Zhao, Song, Chen, Wen, and Jia]{aiscientist_long}
Guoxin Chen, Jie Chen, Lei Chen, Jiale Zhao, Fanzhe Meng, Wayne~Xin Zhao, Ruihua Song, Cheng Chen, Ji{-}Rong Wen, and Kai Jia.
\newblock Toward autonomous long-horizon engineering for {ML} research.
\newblock \emph{CoRR}, abs/2604.13018, 2026{\natexlab{a}}.
\newblock \doi{10.48550/ARXIV.2604.13018}.
\newblock \url{https://doi.org/10.48550/arXiv.2604.13018}.

\bibitem[Chen et~al.(2026{\natexlab{b}})Chen, Mishra, Nam, Meng, Pfister, and Yoon]{mars}
Jiefeng Chen, Bhavana~Dalvi Mishra, Jaehyun Nam, Rui Meng, Tomas Pfister, and Jinsung Yoon.
\newblock {MARS:} modular agent with reflective search for automated {AI} research.
\newblock \emph{CoRR}, abs/2602.02660, 2026{\natexlab{b}}.
\newblock \doi{10.48550/ARXIV.2602.02660}.
\newblock \url{https://doi.org/10.48550/arXiv.2602.02660}.

\bibitem[Chen et~al.(2024)Chen, Chen, Ning, Zhang, Wang, Yu, Li, Liao, Wei, Lu, Dey, Xue, Baker, Burns, Adu{-}Ampratwum, Huang, Ning, Gao, Su, and Sun]{scienceagentbench}
Ziru Chen, Shijie Chen, Yuting Ning, Qianheng Zhang, Boshi Wang, Botao Yu, Yifei Li, Zeyi Liao, Chen Wei, Zitong Lu, Vishal Dey, Mingyi Xue, Frazier~N. Baker, Benjamin Burns, Daniel Adu{-}Ampratwum, Xuhui Huang, Xia Ning, Song Gao, Yu~Su, and Huan Sun.
\newblock Scienceagentbench: Toward rigorous assessment of language agents for data-driven scientific discovery.
\newblock \emph{CoRR}, abs/2410.05080, 2024.
\newblock \doi{10.48550/ARXIV.2410.05080}.
\newblock \url{https://doi.org/10.48550/arXiv.2410.05080}.

\bibitem[Chi et~al.(2026)Chi, Hong, Jiang, Luo, Yang, Zhang, Cao, Fan, He, Hao, Jin, Lei, Liu, Qian, Wang, Wang, Zheng, Zhou, Xiao, Cai, and Na]{frontiereng}
Yizhe Chi, Deyao Hong, Dapeng Jiang, Tianwei Luo, Kaisen Yang, Boshi Zhang, Zhe Cao, Xiaoyan Fan, Bingxiang He, Han Hao, Weiyang Jin, Dianqiao Lei, Qingle Liu, Houde Qian, Bowen Wang, Situ Wang, Youjie Zheng, Yifan Zhou, Calvin Xiao, Eren Cai, and Qinhuai Na.
\newblock Frontier-eng: Benchmarking self-evolving agents on real-world engineering tasks with generative optimization.
\newblock \emph{CoRR}, abs/2604.12290, 2026.
\newblock \doi{10.48550/ARXIV.2604.12290}.
\newblock \url{https://doi.org/10.48550/arXiv.2604.12290}.

\bibitem[Du et~al.(2026)Du, Yang, Zhou, Liu, Lei, Chen, Liu, Wu, Cai, Liu, Zhu, Wang, Zhang, Qian, and Chen]{datamaster}
Yaxin Du, Xiyuan Yang, Zhifan Zhou, Wanxu Liu, Zixing Lei, Zimeng Chen, Fenyi Liu, Haotian Wu, Yuzhu Cai, Zexi Liu, Xinyu Zhu, WenHao Wang, Linfeng Zhang, Chen Qian, and Siheng Chen.
\newblock Datamaster: Data-centric autonomous ai research, 2026.
\newblock \url{https://arxiv.org/abs/2605.10906}.

\bibitem[Gasteiger et~al.(2025)Gasteiger, Khan, Bowman, Mikulik, Perez, and Roger]{gasteiger2025automated}
Johannes Gasteiger, Akbir Khan, Sam Bowman, Vladimir Mikulik, Ethan Perez, and Fabien Roger.
\newblock Automated researchers can subtly sandbag, March 2025.
\newblock \url{https://alignment.anthropic.com/2025/automated-researchers-sandbag/}.

\bibitem[Hu et~al.(2025)Hu, Liu, Yue, Zhang, Liu, Zhu, Lin, Guo, Dou, Xi, Jin, Tan, Yin, Liu, Zhang, Sun, Zhu, Sun, Peng, Cheng, Fan, Guo, Yu, Zhou, Hu, Huo, Wang, Niu, Wang, Yin, Hu, Liao, Li, Wang, Zhou, Liu, Cheng, Zhang, Gui, Pan, Zhang, Torr, Dou, Wen, Huang, Jiang, and Yan]{memorysurvey}
Yuyang Hu, Shichun Liu, Yanwei Yue, Guibin Zhang, Boyang Liu, Fangyi Zhu, Jiahang Lin, Honglin Guo, Shihan Dou, Zhiheng Xi, Senjie Jin, Jiejun Tan, Yanbin Yin, Jiongnan Liu, Zeyu Zhang, Zhongxiang Sun, Yutao Zhu, Hao Sun, Boci Peng, Zhenrong Cheng, Xuanbo Fan, Jiaxin Guo, Xinlei Yu, Zhenhong Zhou, Zewen Hu, Jiahao Huo, Junhao Wang, Yuwei Niu, Yu~Wang, Zhenfei Yin, Xiaobin Hu, Yue Liao, Qiankun Li, Kun Wang, Wangchunshu Zhou, Yixin Liu, Dawei Cheng, Qi~Zhang, Tao Gui, Shirui Pan, Yan Zhang, Philip Torr, Zhicheng Dou, Ji{-}Rong Wen, Xuanjing Huang, Yu{-}Gang Jiang, and Shuicheng Yan.
\newblock Memory in the age of {AI} agents.
\newblock \emph{CoRR}, abs/2512.13564, 2025.
\newblock \doi{10.48550/ARXIV.2512.13564}.
\newblock \url{https://doi.org/10.48550/arXiv.2512.13564}.

\bibitem[Hu et~al.(2026{\natexlab{a}})Hu, Qian, Wang, Liu, Zhao, Li, Liu, and Dou]{agentfugue}
Yuyang Hu, Hongjin Qian, Shuting Wang, Jiongnan Liu, Tong Zhao, Xiaoxi Li, Zheng Liu, and Zhicheng Dou.
\newblock Agentfugue: Agent scaling for long-horizon tasks through collective reasoning, 2026{\natexlab{a}}.
\newblock \url{https://arxiv.org/abs/2605.24486}.

\bibitem[Hu et~al.(2026{\natexlab{b}})Hu, Qian, Wang, Liu, Zhao, Tan, Liu, and Dou]{sam}
Yuyang Hu, Hongjin Qian, Shuting Wang, Jiongnan Liu, Ziliang Zhao, Jiejun Tan, Zheng Liu, and Zhicheng Dou.
\newblock Sam: State-adaptive memory for long-horizon reasoning agent, 2026{\natexlab{b}}.
\newblock \url{https://arxiv.org/abs/2605.24468}.

\bibitem[Huang et~al.(2024)Huang, Vora, Liang, and Leskovec]{mlagentbench}
Qian Huang, Jian Vora, Percy Liang, and Jure Leskovec.
\newblock Mlagentbench: Evaluating language agents on machine learning experimentation.
\newblock In Ruslan Salakhutdinov, Zico Kolter, Katherine~A. Heller, Adrian Weller, Nuria Oliver, Jonathan Scarlett, and Felix Berkenkamp, editors, \emph{Forty-first International Conference on Machine Learning, {ICML} 2024, Vienna, Austria, July 21-27, 2024}, Proceedings of Machine Learning Research, pages 20271--20309. {PMLR} / OpenReview.net, 2024.
\newblock \url{https://proceedings.mlr.press/v235/huang24y.html}.

\bibitem[Jiang et~al.(2025)Jiang, Schmidt, Srikanth, Xu, Kaplan, Jacenko, and Wu]{aide}
Zhengyao Jiang, Dominik Schmidt, Dhruv Srikanth, Dixing Xu, Ian Kaplan, Deniss Jacenko, and Yuxiang Wu.
\newblock {AIDE:} ai-driven exploration in the space of code.
\newblock \emph{CoRR}, abs/2502.13138, 2025.
\newblock \doi{10.48550/ARXIV.2502.13138}.
\newblock \url{https://doi.org/10.48550/arXiv.2502.13138}.

\bibitem[Jordan and contributors(2025)]{nanogptbench}
Keller Jordan and contributors.
\newblock {NanoGPT-Bench}: {NanoGPT} training speedrun benchmark.
\newblock \url{https://github.com/KellerJordan/modded-nanogpt}, 2025.

\bibitem[Karpathy(2026)]{autoresearch}
Andrej Karpathy.
\newblock autoresearch: {AI} agents running research on single-{GPU} nanochat training automatically.
\newblock \url{https://github.com/karpathy/autoresearch}, 2026.

\bibitem[Kwa et~al.(2025)Kwa, West, Becker, Deng, Garcia, Hasin, Jawhar, Kinniment, Rush, von Arx, Bloom, Broadley, Du, Goodrich, Jurkovic, Miles, Nix, Lin, Parikh, Rein, Sato, Wijk, Ziegler, Barnes, and Chan]{metr}
Thomas Kwa, Ben West, Joel Becker, Amy Deng, Katharyn Garcia, Max Hasin, Sami Jawhar, Megan Kinniment, Nate Rush, Sydney von Arx, Ryan Bloom, Thomas Broadley, Haoxing Du, Brian Goodrich, Nikola Jurkovic, Luke~Harold Miles, Seraphina Nix, Tao Lin, Neev Parikh, David Rein, Lucas Jun~Koba Sato, Hjalmar Wijk, Daniel~M. Ziegler, Elizabeth Barnes, and Lawrence Chan.
\newblock Measuring {AI} ability to complete long tasks.
\newblock \emph{CoRR}, abs/2503.14499, 2025.
\newblock \doi{10.48550/ARXIV.2503.14499}.
\newblock \url{https://doi.org/10.48550/arXiv.2503.14499}.

\bibitem[Lee et~al.(2026)Lee, Nair, Zhang, Lee, Khattab, and Finn]{metaharness}
Yoonho Lee, Roshen Nair, Qizheng Zhang, Kangwook Lee, Omar Khattab, and Chelsea Finn.
\newblock Meta-harness: End-to-end optimization of model harnesses.
\newblock \emph{CoRR}, abs/2603.28052, 2026.
\newblock \doi{10.48550/ARXIV.2603.28052}.
\newblock \url{https://doi.org/10.48550/arXiv.2603.28052}.

\bibitem[Li et~al.(2025)Li, Wu, Ge, Chong, Hou, Cao, Ju, Wu, Li, Zhang, Feng, Zhao, Qiu, Yang, Zhang, Zhu, Sun, Sun, Yan, Liu, Yin, and Shen]{famouagent}
Annan Li, Chufan Wu, Zengle Ge, Yee~Hin Chong, Zhinan Hou, Lizhe Cao, Cheng Ju, Jianmin Wu, Huaiming Li, Haobo Zhang, Shenghao Feng, Mo~Zhao, Fengzhi Qiu, Rui Yang, Mengmeng Zhang, Wenyi Zhu, Yingying Sun, Quan Sun, Shunhao Yan, Danyu Liu, Dawei Yin, and Dou Shen.
\newblock The {FM} agent.
\newblock \emph{CoRR}, abs/2510.26144, 2025.
\newblock \doi{10.48550/ARXIV.2510.26144}.
\newblock \url{https://doi.org/10.48550/arXiv.2510.26144}.

\bibitem[Lin et~al.(2026)Lin, Liu, Pan, Lin, Dou, Huang, Yan, Han, and Gui]{ahe}
Jiahang Lin, Shichun Liu, Chengjun Pan, Lizhi Lin, Shihan Dou, Xuanjing Huang, Hang Yan, Zhenhua Han, and Tao Gui.
\newblock Agentic harness engineering: Observability-driven automatic evolution of coding-agent harnesses.
\newblock \emph{CoRR}, abs/2604.25850, 2026.
\newblock \doi{10.48550/ARXIV.2604.25850}.
\newblock \url{https://doi.org/10.48550/arXiv.2604.25850}.

\bibitem[Liu et~al.(2025{\natexlab{a}})Liu, Zhu, Bai, He, Liao, Que, Wang, Zhang, Zhang, Zhang, Zhang, Chen, Guo, Li, Liu, Shan, Song, Tian, Wu, Zhou, Zhu, Feng, Gao, He, Li, Liu, Meng, Su, Tan, Wang, Yang, Ye, Zheng, Zhou, Huang, Li, and Zhang]{longcontextsurvey}
Jiaheng Liu, Dawei Zhu, Zhiqi Bai, Yancheng He, Huanxuan Liao, Haoran Que, Zekun Wang, Chenchen Zhang, Ge~Zhang, Jiebin Zhang, Yuanxing Zhang, Zhuo Chen, Hangyu Guo, Shilong Li, Ziqiang Liu, Yong Shan, Yifan Song, Jiayi Tian, Wenhao Wu, Zhejian Zhou, Ruijie Zhu, Junlan Feng, Yang Gao, Shizhu He, Zhoujun Li, Tianyu Liu, Fanyu Meng, Wenbo Su, Yingshui Tan, Zili Wang, Jian Yang, Wei Ye, Bo~Zheng, Wangchunshu Zhou, Wenhao Huang, Sujian Li, and Zhaoxiang Zhang.
\newblock A comprehensive survey on long context language modeling.
\newblock \emph{CoRR}, abs/2503.17407, 2025{\natexlab{a}}.
\newblock \doi{10.48550/ARXIV.2503.17407}.
\newblock \url{https://doi.org/10.48550/arXiv.2503.17407}.

\bibitem[Liu et~al.(2025{\natexlab{b}})Liu, Cai, Zhu, Zheng, Chen, Wen, Wang, E, and Chen]{mlmaster}
Zexi Liu, Yuzhu Cai, Xinyu Zhu, Yujie Zheng, Runkun Chen, Ying Wen, Yanfeng Wang, Weinan E, and Siheng Chen.
\newblock Ml-master: Towards ai-for-ai via integration of exploration and reasoning.
\newblock \emph{CoRR}, abs/2506.16499, 2025{\natexlab{b}}.
\newblock \doi{10.48550/ARXIV.2506.16499}.
\newblock \url{https://doi.org/10.48550/arXiv.2506.16499}.

\bibitem[Lou et~al.(2026)Lou, L{\'{a}}zaro{-}Gredilla, Dedieu, Wendelken, Lehrach, and Murphy]{autoharness}
Xinghua Lou, Miguel L{\'{a}}zaro{-}Gredilla, Antoine Dedieu, Carter Wendelken, Wolfgang Lehrach, and Kevin~P. Murphy.
\newblock Autoharness: improving {LLM} agents by automatically synthesizing a code harness.
\newblock \emph{CoRR}, abs/2603.03329, 2026.
\newblock \doi{10.48550/ARXIV.2603.03329}.
\newblock \url{https://doi.org/10.48550/arXiv.2603.03329}.

\bibitem[Lu et~al.(2024)Lu, Lu, Lange, Foerster, Clune, and Ha]{aiscientist}
Chris Lu, Cong Lu, Robert~Tjarko Lange, Jakob~N. Foerster, Jeff Clune, and David Ha.
\newblock The {AI} scientist: Towards fully automated open-ended scientific discovery.
\newblock \emph{CoRR}, abs/2408.06292, 2024.
\newblock \doi{10.48550/ARXIV.2408.06292}.
\newblock \url{https://doi.org/10.48550/arXiv.2408.06292}.

\bibitem[Merrill et~al.(2026)Merrill, Shaw, Carlini, Li, Raj, Bercovich, Shi, Shin, Walshe, Buchanan, Shen, Ye, Lin, Poulos, Wang, Nezhurina, Jitsev, Lu, Menis{-}Mastromichalakis, Xu, Chen, Liu, Zhang, Chen, Kashyap, Uslu, Li, Wu, Yan, Bian, Sharma, Sun, Dillmann, Anand, Lanpouthakoun, Koopah, Hu, Guha, Dreiman, Zhu, Krauth, Zhong, Muennighoff, Amanfu, Tan, Pimpalgaonkar, Aggarwal, Lin, Lan, Zhao, Liang, Wang, Wang, Zhou, Heineman, Liu, Trivedi, Yang, Lin, Shetty, Yang, Omi, Raoof, Li, Zhuo, Lin, Dai, Wang, Chai, Zhou, Wahdany, She, Hu, Dong, Zhu, Cui, Saiyed, Kolbeinsson, Hu, Rytting, Marten, Wang, Dimakis, Konwinski, and Schmidt]{terminalbench}
Mike~A. Merrill, Alexander~Glenn Shaw, Nicholas Carlini, Boxuan Li, Harsh Raj, Ivan Bercovich, Lin Shi, Jeong~Yeon Shin, Thomas Walshe, Estefany~Kelly Buchanan, Junhong Shen, Guanghao Ye, Haowei Lin, Jason Poulos, Maoyu Wang, Marianna Nezhurina, Jenia Jitsev, Di~Lu, Orfeas Menis{-}Mastromichalakis, Zhiwei Xu, Zizhao Chen, Yue Liu, Robert Zhang, Leon~Liangyu Chen, Anurag Kashyap, Jan{-}Lucas Uslu, Jeffrey Li, Jianbo Wu, Minghao Yan, Song Bian, Vedang Sharma, Ke~Sun, Steven Dillmann, Akshay Anand, Andrew Lanpouthakoun, Bardia Koopah, Changran Hu, Etash~Kumar Guha, Gabriel H.~S. Dreiman, Jiacheng Zhu, Karl Krauth, Li~Zhong, Niklas Muennighoff, Robert Amanfu, Shangyin Tan, Shreyas Pimpalgaonkar, Tushar Aggarwal, Xiangning Lin, Xin Lan, Xuandong Zhao, Yiqing Liang, Yuanli Wang, Zilong Wang, Changzhi Zhou, David Heineman, Hange Liu, Harsh Trivedi, John Yang, Junhong Lin, Manish Shetty, Michael Yang, Nabil Omi, Negin Raoof, Shanda Li, Terry~Yue Zhuo, Wuwei Lin, Yiwei Dai, Yuxin Wang, Wenhao Chai, Shang Zhou, Dariush
  Wahdany, Ziyu She, Jiaming Hu, Zhikang Dong, Yuxuan Zhu, Sasha Cui, Ahson Saiyed, Arinbj{\"{o}}rn Kolbeinsson, Jesse Hu, Christopher~Michael Rytting, Ryan Marten, Yixin Wang, Alex Dimakis, Andy Konwinski, and Ludwig Schmidt.
\newblock Terminal-bench: Benchmarking agents on hard, realistic tasks in command line interfaces.
\newblock \emph{CoRR}, abs/2601.11868, 2026.
\newblock \doi{10.48550/ARXIV.2601.11868}.
\newblock \url{https://doi.org/10.48550/arXiv.2601.11868}.

\bibitem[Nadafian et~al.(2026)Nadafian, Mohammadshahi, and Yazdani]{kapso}
Alireza Nadafian, Alireza Mohammadshahi, and Majid Yazdani.
\newblock {KAPSO:} {A} knowledge-grounded framework for autonomous program synthesis and optimization.
\newblock \emph{CoRR}, abs/2601.21526, 2026.
\newblock \doi{10.48550/ARXIV.2601.21526}.
\newblock \url{https://doi.org/10.48550/arXiv.2601.21526}.

\bibitem[Novikov et~al.(2025)Novikov, Vu, Eisenberger, Dupont, Huang, Wagner, Shirobokov, Kozlovskii, Ruiz, Mehrabian, Kumar, See, Chaudhuri, Holland, Davies, Nowozin, Kohli, and Balog]{alphaevolve}
Alexander Novikov, Ng{\^{a}}n Vu, Marvin Eisenberger, Emilien Dupont, Po{-}Sen Huang, Adam~Zsolt Wagner, Sergey Shirobokov, Borislav Kozlovskii, Francisco J.~R. Ruiz, Abbas Mehrabian, M.~Pawan Kumar, Abigail See, Swarat Chaudhuri, George Holland, Alex Davies, Sebastian Nowozin, Pushmeet Kohli, and Matej Balog.
\newblock Alphaevolve: {A} coding agent for scientific and algorithmic discovery.
\newblock \emph{CoRR}, abs/2506.13131, 2025.
\newblock \doi{10.48550/ARXIV.2506.13131}.
\newblock \url{https://doi.org/10.48550/arXiv.2506.13131}.

\bibitem[{OpenAI}(2025)]{codex}
{OpenAI}.
\newblock {Codex CLI}.
\newblock \url{https://github.com/openai/codex}, 2025.
\newblock Lightweight coding agent that runs locally on a user's computer. Accessed: 2026-06-02.

\bibitem[Park et~al.(2023)Park, O'Brien, Cai, Morris, Liang, and Bernstein]{generativeagent}
Joon~Sung Park, Joseph~C. O'Brien, Carrie~Jun Cai, Meredith~Ringel Morris, Percy Liang, and Michael~S. Bernstein.
\newblock Generative agents: Interactive simulacra of human behavior.
\newblock In Sean Follmer, Jeff Han, J{\"{u}}rgen Steimle, and Nathalie~Henry Riche, editors, \emph{Proceedings of the 36th Annual {ACM} Symposium on User Interface Software and Technology, {UIST} 2023, San Francisco, CA, USA, 29 October 2023- 1 November 2023}, pages 2:1--2:22. {ACM}, 2023.
\newblock \doi{10.1145/3586183.3606763}.
\newblock \url{https://doi.org/10.1145/3586183.3606763}.

\bibitem[Press et~al.(2025)Press, Amos, Zhao, Wu, Ainsworth, Krupke, Kidger, Sajed, Stellato, Park, Bosch, Meril, Steppi, Zharmagambetov, Zhang, Perez{-}Pineiro, Mercurio, Zhan, Abramovich, Lieret, Zhang, Huang, Bethge, and Press]{algotune}
Ori Press, Brandon Amos, Haoyu Zhao, Yikai Wu, Samuel~K. Ainsworth, Dominik Krupke, Patrick Kidger, Touqir Sajed, Bartolomeo Stellato, Jisun Park, Nathanael Bosch, Eli Meril, Albert Steppi, Arman Zharmagambetov, Fangzhao Zhang, David Perez{-}Pineiro, Alberto Mercurio, Ni~Zhan, Talor Abramovich, Kilian Lieret, Hanlin Zhang, Shirley Huang, Matthias Bethge, and Ofir Press.
\newblock Algotune: Can language models speed up general-purpose numerical programs?
\newblock \emph{CoRR}, abs/2507.15887, 2025.
\newblock \doi{10.48550/ARXIV.2507.15887}.
\newblock \url{https://doi.org/10.48550/arXiv.2507.15887}.

\bibitem[Qiang et~al.(2025)Qiang, Zhuang, Li, K, Zhang, Li, Wong, Yang, Liang, Zhang, and Dai]{mledojo}
Rushi Qiang, Yuchen Zhuang, Yinghao Li, Dingu Sagar~V. K, Rongzhi Zhang, Changhao Li, Ian~Shu{-}Hei Wong, Sherry Yang, Percy Liang, Chao Zhang, and Bo~Dai.
\newblock Mle-dojo: Interactive environments for empowering {LLM} agents in machine learning engineering.
\newblock \emph{CoRR}, abs/2505.07782, 2025.
\newblock \doi{10.48550/ARXIV.2505.07782}.
\newblock \url{https://doi.org/10.48550/arXiv.2505.07782}.

\bibitem[Qin et~al.(2024)Qin, Liang, Ye, Zhu, Yan, Lu, Lin, Cong, Tang, Qian, Zhao, Hong, Tian, Xie, Zhou, Gerstein, Li, Liu, and Sun]{toolllm}
Yujia Qin, Shihao Liang, Yining Ye, Kunlun Zhu, Lan Yan, Yaxi Lu, Yankai Lin, Xin Cong, Xiangru Tang, Bill Qian, Sihan Zhao, Lauren Hong, Runchu Tian, Ruobing Xie, Jie Zhou, Mark Gerstein, Dahai Li, Zhiyuan Liu, and Maosong Sun.
\newblock Toolllm: Facilitating large language models to master 16000+ real-world apis.
\newblock In \emph{The Twelfth International Conference on Learning Representations, {ICLR} 2024, Vienna, Austria, May 7-11, 2024}. OpenReview.net, 2024.
\newblock \url{https://openreview.net/forum?id=dHng2O0Jjr}.

\bibitem[Rank et~al.(2026)Rank, Bhatnagar, Prabhu, Eisenberg, Nguyen, Bethge, and Andriushchenko]{posttrainbench}
Ben Rank, Hardik Bhatnagar, Ameya Prabhu, Shira Eisenberg, Karina Nguyen, Matthias Bethge, and Maksym Andriushchenko.
\newblock Posttrainbench: Can {LLM} agents automate {LLM} post-training?
\newblock \emph{CoRR}, abs/2603.08640, 2026.
\newblock \doi{10.48550/ARXIV.2603.08640}.
\newblock \url{https://doi.org/10.48550/arXiv.2603.08640}.

\bibitem[Rein et~al.(2025)Rein, Becker, Deng, Nix, Canal, O'Connel, Arnott, Bloom, Broadley, Garcia, Goodrich, Hasin, Jawhar, Kinniment, Kwa, Lajko, Rush, Sato, von Arx, West, Chan, and Barnes]{hcast}
David Rein, Joel Becker, Amy Deng, Seraphina Nix, Chris Canal, Daniel O'Connel, Pip Arnott, Ryan Bloom, Thomas Broadley, Katharyn Garcia, Brian Goodrich, Max Hasin, Sami Jawhar, Megan Kinniment, Thomas Kwa, Aron Lajko, Nate Rush, Lucas Jun~Koba Sato, Sydney von Arx, Ben West, Lawrence Chan, and Elizabeth Barnes.
\newblock {HCAST:} human-calibrated autonomy software tasks.
\newblock \emph{CoRR}, abs/2503.17354, 2025.
\newblock \doi{10.48550/ARXIV.2503.17354}.
\newblock \url{https://doi.org/10.48550/arXiv.2503.17354}.

\bibitem[Romera{-}Paredes et~al.(2024)Romera{-}Paredes, Barekatain, Novikov, Balog, Kumar, Dupont, Ruiz, Ellenberg, Wang, Fawzi, Kohli, and Fawzi]{funsearch}
Bernardino Romera{-}Paredes, Mohammadamin Barekatain, Alexander Novikov, Matej Balog, M.~Pawan Kumar, Emilien Dupont, Francisco J.~R. Ruiz, Jordan~S. Ellenberg, Pengming Wang, Omar Fawzi, Pushmeet Kohli, and Alhussein Fawzi.
\newblock Mathematical discoveries from program search with large language models.
\newblock \emph{Nat.}, 625\penalty0 (7995):\penalty0 468--475, 2024.
\newblock \doi{10.1038/S41586-023-06924-6}.
\newblock \url{https://doi.org/10.1038/s41586-023-06924-6}.

\bibitem[Schick et~al.(2023)Schick, Dwivedi{-}Yu, Dess{\`{\i}}, Raileanu, Lomeli, Hambro, Zettlemoyer, Cancedda, and Scialom]{toolformer}
Timo Schick, Jane Dwivedi{-}Yu, Roberto Dess{\`{\i}}, Roberta Raileanu, Maria Lomeli, Eric Hambro, Luke Zettlemoyer, Nicola Cancedda, and Thomas Scialom.
\newblock Toolformer: Language models can teach themselves to use tools.
\newblock In Alice Oh, Tristan Naumann, Amir Globerson, Kate Saenko, Moritz Hardt, and Sergey Levine, editors, \emph{Advances in Neural Information Processing Systems 36: Annual Conference on Neural Information Processing Systems 2023, NeurIPS 2023, New Orleans, LA, USA, December 10 - 16, 2023}, 2023.
\newblock \url{http://papers.nips.cc/paper\_files/paper/2023/hash/d842425e4bf79ba039352da0f658a906-Abstract-Conference.html}.

\bibitem[Schmidgall et~al.(2025)Schmidgall, Su, Wang, Sun, Wu, Yu, Liu, Liu, and Barsoum]{agentlaboratory}
Samuel Schmidgall, Yusheng Su, Ze~Wang, Ximeng Sun, Jialian Wu, Xiaodong Yu, Jiang Liu, Zicheng Liu, and Emad Barsoum.
\newblock Agent laboratory: Using {LLM} agents as research assistants.
\newblock \emph{CoRR}, abs/2501.04227, 2025.
\newblock \doi{10.48550/ARXIV.2501.04227}.
\newblock \url{https://doi.org/10.48550/arXiv.2501.04227}.

\bibitem[Shinn et~al.(2023)Shinn, Cassano, Gopinath, Narasimhan, and Yao]{reflexion}
Noah Shinn, Federico Cassano, Ashwin Gopinath, Karthik Narasimhan, and Shunyu Yao.
\newblock Reflexion: language agents with verbal reinforcement learning.
\newblock In Alice Oh, Tristan Naumann, Amir Globerson, Kate Saenko, Moritz Hardt, and Sergey Levine, editors, \emph{Advances in Neural Information Processing Systems 36: Annual Conference on Neural Information Processing Systems 2023, NeurIPS 2023, New Orleans, LA, USA, December 10 - 16, 2023}, 2023.
\newblock \url{http://papers.nips.cc/paper\_files/paper/2023/hash/1b44b878bb782e6954cd888628510e90-Abstract-Conference.html}.

\bibitem[Sinha et~al.(2025)Sinha, Arun, Goel, Staab, and Geiping]{longhorizonexecution}
Akshit Sinha, Arvindh Arun, Shashwat Goel, Steffen Staab, and Jonas Geiping.
\newblock The illusion of diminishing returns: Measuring long horizon execution in llms.
\newblock \emph{CoRR}, abs/2509.09677, 2025.
\newblock \doi{10.48550/ARXIV.2509.09677}.
\newblock \url{https://doi.org/10.48550/arXiv.2509.09677}.

\bibitem[Starace et~al.(2025)Starace, Jaffe, Sherburn, Aung, Chan, Maksin, Dias, Mays, Kinsella, Thompson, Heidecke, Glaese, and Patwardhan]{paperbench}
Giulio Starace, Oliver Jaffe, Dane Sherburn, James Aung, Jun~Shern Chan, Leon Maksin, Rachel Dias, Evan Mays, Benjamin Kinsella, Wyatt Thompson, Johannes Heidecke, Amelia Glaese, and Tejal Patwardhan.
\newblock Paperbench: Evaluating ai's ability to replicate {AI} research.
\newblock \emph{CoRR}, abs/2504.01848, 2025.
\newblock \doi{10.48550/ARXIV.2504.01848}.
\newblock \url{https://doi.org/10.48550/arXiv.2504.01848}.

\bibitem[Team(2026)]{internagent}
InternScience Team.
\newblock Internagent-1.5: {A} unified agentic framework for long-horizon autonomous scientific discovery.
\newblock \emph{CoRR}, abs/2602.08990, 2026.
\newblock \doi{10.48550/ARXIV.2602.08990}.
\newblock \url{https://doi.org/10.48550/arXiv.2602.08990}.

\bibitem[Tie et~al.(2025)Tie, Zhou, and Sun]{aiscientistsurvey}
Guiyao Tie, Pan Zhou, and Lichao Sun.
\newblock A survey of {AI} scientists.
\newblock \emph{CoRR}, abs/2510.23045, 2025.
\newblock \doi{10.48550/ARXIV.2510.23045}.
\newblock \url{https://doi.org/10.48550/arXiv.2510.23045}.

\bibitem[Toledo et~al.(2025)Toledo, Hambardzumyan, Josifoski, Hazra, Baldwin, Audran{-}Reiss, Kuchnik, Magka, Jiang, Lupidi, Lupu, Raileanu, Niu, Shavrina, Gagnon{-}Audet, Shvartsman, Sodhani, Miller, Charnalia, Dunfield, Wu, Stenetorp, Cancedda, Foerster, and Bachrach]{airadojo}
Edan Toledo, Karen Hambardzumyan, Martin Josifoski, Rishi Hazra, Nicolas~Mario Baldwin, Alexis Audran{-}Reiss, Michael Kuchnik, Despoina Magka, Minqi Jiang, Alisia~Maria Lupidi, Andrei Lupu, Roberta Raileanu, Kelvin Niu, Tatiana Shavrina, Jean{-}Christophe Gagnon{-}Audet, Michael Shvartsman, Shagun Sodhani, Alexander~H. Miller, Abhishek Charnalia, Derek Dunfield, Carole{-}Jean Wu, Pontus Stenetorp, Nicola Cancedda, Jakob~Nicolaus Foerster, and Yoram Bachrach.
\newblock {AI} research agents for machine learning: Search, exploration, and generalization in mle-bench.
\newblock \emph{CoRR}, abs/2507.02554, 2025.
\newblock \doi{10.48550/ARXIV.2507.02554}.
\newblock \url{https://doi.org/10.48550/arXiv.2507.02554}.

\bibitem[Wan et~al.(2025)Wan, Dai, Wang, Li, Wang, Mao, Lan, and Xiao]{loongflow}
Chunhui Wan, Xunan Dai, Zhuo Wang, Minglei Li, Yanpeng Wang, Yinan Mao, Yu~Lan, and Zhiwen Xiao.
\newblock Loongflow: Directed evolutionary search via a cognitive plan-execute-summarize paradigm.
\newblock \emph{CoRR}, abs/2512.24077, 2025.
\newblock \doi{10.48550/ARXIV.2512.24077}.
\newblock \url{https://doi.org/10.48550/arXiv.2512.24077}.

\bibitem[Wang et~al.(2026)Wang, Lin, Hu, Jiao, Chowdhury, Chang, and Patwardhan]{frontierscience}
Miles Wang, Robi Lin, Kat Hu, Joy Jiao, Neil Chowdhury, Ethan Chang, and Tejal Patwardhan.
\newblock Frontierscience: Evaluating ai's ability to perform expert-level scientific tasks.
\newblock \emph{CoRR}, abs/2601.21165, 2026.
\newblock \doi{10.48550/ARXIV.2601.21165}.
\newblock \url{https://doi.org/10.48550/arXiv.2601.21165}.

\bibitem[Wang et~al.(2025)Wang, Li, Song, Xu, Tang, Zhuge, Pan, Song, Li, Singh, Tran, Li, Ma, Zheng, Qian, Shao, Muennighoff, Zhang, Hui, Lin, and et~al.]{openhands}
Xingyao Wang, Boxuan Li, Yufan Song, Frank~F. Xu, Xiangru Tang, Mingchen Zhuge, Jiayi Pan, Yueqi Song, Bowen Li, Jaskirat Singh, Hoang~H. Tran, Fuqiang Li, Ren Ma, Mingzhang Zheng, Bill Qian, Yanjun Shao, Niklas Muennighoff, Yizhe Zhang, Binyuan Hui, Junyang Lin, and et~al.
\newblock Openhands: An open platform for {AI} software developers as generalist agents.
\newblock In \emph{The Thirteenth International Conference on Learning Representations, {ICLR} 2025, Singapore, April 24-28, 2025}. OpenReview.net, 2025.
\newblock \url{https://openreview.net/forum?id=OJd3ayDDoF}.

\bibitem[Wei et~al.(2025)Wei, Sun, Papay, McKinney, Han, Fulford, Chung, Passos, Fedus, and Glaese]{browsecomp}
Jason Wei, Zhiqing Sun, Spencer Papay, Scott McKinney, Jeffrey Han, Isa Fulford, Hyung~Won Chung, Alex~Tachard Passos, William Fedus, and Amelia Glaese.
\newblock Browsecomp: {A} simple yet challenging benchmark for browsing agents.
\newblock \emph{CoRR}, abs/2504.12516, 2025.
\newblock \doi{10.48550/ARXIV.2504.12516}.
\newblock \url{https://doi.org/10.48550/arXiv.2504.12516}.

\bibitem[Wijk et~al.(2024)Wijk, Lin, Becker, Jawhar, Parikh, Broadley, Chan, Chen, Clymer, Dhyani, Ericheva, Garcia, Goodrich, Jurkovic, Kinniment, Lajko, Nix, Sato, Saunders, Taran, West, and Barnes]{rebench}
Hjalmar Wijk, Tao Lin, Joel Becker, Sami Jawhar, Neev Parikh, Thomas Broadley, Lawrence Chan, Michael Chen, Joshua Clymer, Jai Dhyani, Elena Ericheva, Katharyn Garcia, Brian Goodrich, Nikola Jurkovic, Megan Kinniment, Aron Lajko, Seraphina Nix, Lucas Sato, William Saunders, Maksym Taran, Ben West, and Elizabeth Barnes.
\newblock Re-bench: Evaluating frontier {AI} r{\&}d capabilities of language model agents against human experts.
\newblock \emph{CoRR}, abs/2411.15114, 2024.
\newblock \doi{10.48550/ARXIV.2411.15114}.
\newblock \url{https://doi.org/10.48550/arXiv.2411.15114}.

\bibitem[Yamada et~al.(2025)Yamada, Lange, Lu, Hu, Lu, Foerster, Clune, and Ha]{aiscientistv2}
Yutaro Yamada, Robert~Tjarko Lange, Cong Lu, Shengran Hu, Chris Lu, Jakob~N. Foerster, Jeff Clune, and David Ha.
\newblock The {AI} scientist-v2: Workshop-level automated scientific discovery via agentic tree search.
\newblock \emph{CoRR}, abs/2504.08066, 2025.
\newblock \doi{10.48550/ARXIV.2504.08066}.
\newblock \url{https://doi.org/10.48550/arXiv.2504.08066}.

\bibitem[Yang et~al.(2025)Yang, Yang, Fang, Xian, Li, Wang, Xu, Pan, Hong, Liu, Shen, Chen, and Bian]{rdagent}
Xu~Yang, Xiao Yang, Shikai Fang, Bowen Xian, Yuante Li, Jian Wang, Minrui Xu, Haoran Pan, Xinpeng Hong, Weiqing Liu, Yelong Shen, Weizhu Chen, and Jiang Bian.
\newblock R{\&}d-agent: Automating data-driven {AI} solution building through llm-powered automated research, development, and evolution.
\newblock \emph{CoRR}, abs/2505.14738, 2025.
\newblock \doi{10.48550/ARXIV.2505.14738}.
\newblock \url{https://doi.org/10.48550/arXiv.2505.14738}.

\bibitem[Yao et~al.(2023{\natexlab{a}})Yao, Yu, Zhao, Shafran, Griffiths, Cao, and Narasimhan]{treeofthoughts}
Shunyu Yao, Dian Yu, Jeffrey Zhao, Izhak Shafran, Thomas~L. Griffiths, Yuan Cao, and Karthik Narasimhan.
\newblock Tree of thoughts: Deliberate problem solving with large language models.
\newblock \emph{CoRR}, abs/2305.10601, 2023{\natexlab{a}}.
\newblock \doi{10.48550/ARXIV.2305.10601}.
\newblock \url{https://doi.org/10.48550/arXiv.2305.10601}.

\bibitem[Yao et~al.(2023{\natexlab{b}})Yao, Zhao, Yu, Du, Shafran, Narasimhan, and Cao]{react}
Shunyu Yao, Jeffrey Zhao, Dian Yu, Nan Du, Izhak Shafran, Karthik~R. Narasimhan, and Yuan Cao.
\newblock React: Synergizing reasoning and acting in language models.
\newblock In \emph{The Eleventh International Conference on Learning Representations, {ICLR} 2023, Kigali, Rwanda, May 1-5, 2023}. OpenReview.net, 2023{\natexlab{b}}.
\newblock \url{https://openreview.net/forum?id=WE\_vluYUL-X}.

\bibitem[Zhang et~al.(2025{\natexlab{a}})Zhang, Hu, Lu, Lange, and Clune]{darwinmachine}
Jenny Zhang, Shengran Hu, Cong Lu, Robert~T. Lange, and Jeff Clune.
\newblock Darwin godel machine: Open-ended evolution of self-improving agents.
\newblock \emph{CoRR}, abs/2505.22954, 2025{\natexlab{a}}.
\newblock \doi{10.48550/ARXIV.2505.22954}.
\newblock \url{https://doi.org/10.48550/arXiv.2505.22954}.

\bibitem[Zhang et~al.(2025{\natexlab{b}})Zhang, Xiang, Yu, Teng, Chen, Chen, Zhuge, Cheng, Hong, Wang, Zheng, Liu, Luo, and Wu]{aflow}
Jiayi Zhang, Jinyu Xiang, Zhaoyang Yu, Fengwei Teng, Xionghui Chen, Jiaqi Chen, Mingchen Zhuge, Xin Cheng, Sirui Hong, Jinlin Wang, Bingnan Zheng, Bang Liu, Yuyu Luo, and Chenglin Wu.
\newblock Aflow: Automating agentic workflow generation.
\newblock In \emph{The Thirteenth International Conference on Learning Representations, {ICLR} 2025, Singapore, April 24-28, 2025}. OpenReview.net, 2025{\natexlab{b}}.
\newblock \url{https://openreview.net/forum?id=z5uVAKwmjf}.

\bibitem[Zhang et~al.(2025{\natexlab{c}})Zhang, Hu, Upasani, Ma, Hong, Kamanuru, Rainton, Wu, Ji, Li, Thakker, Zou, and Olukotun]{ace}
Qizheng Zhang, Changran Hu, Shubhangi Upasani, Boyuan Ma, Fenglu Hong, Vamsidhar Kamanuru, Jay Rainton, Chen Wu, Mengmeng Ji, Hanchen Li, Urmish Thakker, James Zou, and Kunle Olukotun.
\newblock Agentic context engineering: Evolving contexts for self-improving language models.
\newblock \emph{CoRR}, abs/2510.04618, 2025{\natexlab{c}}.
\newblock \doi{10.48550/ARXIV.2510.04618}.
\newblock \url{https://doi.org/10.48550/arXiv.2510.04618}.

\bibitem[Zhang et~al.(2026)Zhang, Qin, Cao, Zhang, and Xie]{aibuildai}
Ruiyi Zhang, Peijia Qin, Qi~Cao, Li~Zhang, and Pengtao Xie.
\newblock Aibuildai: An {AI} agent for automatically building {AI} models.
\newblock \emph{CoRR}, abs/2604.14455, 2026.
\newblock \doi{10.48550/ARXIV.2604.14455}.
\newblock \url{https://doi.org/10.48550/arXiv.2604.14455}.

\bibitem[Zhao et~al.(2025)Zhao, Magka, Jiang, Li, Raileanu, Shavrina, Gagnon{-}Audet, Niu, Sodhani, Shvartsman, Lupu, Lupidi, Toledo, Hambardzumyan, Josifoski, Foster, Cipolina{-}Kun, Charnalia, Dunfield, Miller, Aodha, Foerster, and Bachrach]{nanogptspeedrunning}
Bingchen Zhao, Despoina Magka, Minqi Jiang, Xian Li, Roberta Raileanu, Tatiana Shavrina, Jean{-}Christophe Gagnon{-}Audet, Kelvin Niu, Shagun Sodhani, Michael Shvartsman, Andrei Lupu, Alisia~Maria Lupidi, Edan Toledo, Karen Hambardzumyan, Martin Josifoski, Thomas Foster, Lucia Cipolina{-}Kun, Abhishek Charnalia, Derek Dunfield, Alexander~H. Miller, Oisin~Mac Aodha, Jakob~N. Foerster, and Yoram Bachrach.
\newblock The automated {LLM} speedrunning benchmark: Reproducing nanogpt improvements.
\newblock \emph{CoRR}, abs/2506.22419, 2025.
\newblock \doi{10.48550/ARXIV.2506.22419}.
\newblock \url{https://doi.org/10.48550/arXiv.2506.22419}.

\bibitem[Zhou et~al.(2023)Zhou, Yan, Shlapentokh{-}Rothman, Wang, and Wang]{lats}
Andy Zhou, Kai Yan, Michal Shlapentokh{-}Rothman, Haohan Wang, and Yu{-}Xiong Wang.
\newblock Language agent tree search unifies reasoning acting and planning in language models.
\newblock \emph{CoRR}, abs/2310.04406, 2023.
\newblock \doi{10.48550/ARXIV.2310.04406}.
\newblock \url{https://doi.org/10.48550/arXiv.2310.04406}.

\bibitem[Zhu et~al.(2026)Zhu, Cai, Liu, Zheng, Wang, Ye, Chen, Wang, Wang, Zhang, Zhang, E, Jin, Chen, and Wang]{mlmaster2}
Xinyu Zhu, Yuzhu Cai, Zexi Liu, Bingyang Zheng, Cheng Wang, Rui Ye, Jiaao Chen, Hanrui Wang, Wei{-}Chen Wang, Yuzhi Zhang, Linfeng Zhang, Weinan E, Di~Jin, Siheng Chen, and Yanfeng Wang.
\newblock Toward ultra-long-horizon agentic science: Cognitive accumulation for machine learning engineering.
\newblock \emph{CoRR}, abs/2601.10402, 2026.
\newblock \doi{10.48550/ARXIV.2601.10402}.
\newblock \url{https://doi.org/10.48550/arXiv.2601.10402}.

\end{thebibliography}
\end{document}